\documentclass[sigconf]{acmart}

\AtBeginDocument{%
  }

\setcopyright{acmlicensed}
\copyrightyear{2018}
\acmYear{2018}
\acmDOI{XXXXXXX.XXXXXXX}

\acmConference[Conference acronym 'XX]{Make sure to enter the correct
  conference title from your rights confirmation emai}{June 03--05,
  2018}{Woodstock, NY}
\acmISBN{978-1-4503-XXXX-X/18/06}

\usepackage{enumitem}
\usepackage{amsmath}
\usepackage{multirow}
\usepackage{graphicx}
\usepackage{subfigure}
\usepackage{wrapfig}
\usepackage{caption}
\usepackage{float}

\begin{document}

\title{SenWave: A Fine-Grained Multi-Language Sentiment Analysis Dataset Sourced from COVID-19 Tweets}


\author{Qiang Yang}
\orcid{0000-0002-3211-5369}
\affiliation{%
  \institution{King Abdullah University of Science and Technology}
  \city{Jeddah}
  \country{Saudi Arabia}
}
\email{qiang.yang@kaust.edu.sa}

\author{Xiuying Chen}
\affiliation{%
  \institution{King Abdullah University of Science and Technology}
  \city{Jeddah}
  \country{Saudi Arabia}}
\email{xiuying.chen@kaust.edu.sa}

\author{Changsheng Ma}
\affiliation{%
  \institution{King Abdullah University of Science and Technology}
  \city{Jeddah}
  \country{Saudi Arabia}}
\email{changsheng.ma@kaust.edu.sa}

\author{Rui Yin}
\affiliation{%
  \institution{University Of Florida}
  \city{Gainesville}
  \country{United States}}
\email{ruiyin@ufl.edu}

\author{Xin Gao}
\affiliation{%
  \institution{King Abdullah University of Science and Technology}
  \city{Jeddah}
  \country{Saudi Arabia}}
\email{xin.gao@kaust.edu.sa}

\author{Xiangliang Zhang}
\authornote{Corresponding author.}
\authornote{Dr. Xiangliang Zhang is secondly affiliated with King Abdullah University of Science and Technology, Saudi Arabia.}
\affiliation{%
  \institution{University of Notre Dame}
  \city{Indiana}
  \country{United States}}
\email{xzhang33@nd.edu}

\renewcommand{\shortauthors}{Qiang Yang et al.}

\begin{abstract}
The global impact of the COVID-19 pandemic has highlighted the need for a comprehensive understanding of public sentiment and reactions. Despite the availability of numerous public datasets on COVID-19, some reaching volumes of up to 100 billion data points, challenges persist regarding the availability of labeled data and the presence of coarse-grained or inappropriate sentiment labels. In this paper, we introduce SenWave, a novel fine-grained multi-language sentiment analysis dataset specifically designed for analyzing COVID-19 tweets, featuring ten sentiment categories across five languages. The dataset comprises 10,000 annotated tweets each in English and Arabic, along with 30,000 translated tweets in Spanish, French, and Italian, derived from English tweets. Additionally, it includes over 105 million unlabeled tweets collected during various COVID-19 waves.
To enable accurate fine-grained sentiment classification, we fine-tuned pre-trained transformer-based language models using the labeled tweets. Our study provides an in-depth analysis of the evolving emotional landscape across languages, countries, and topics, revealing significant insights over time. Furthermore, we assess the compatibility of our dataset with ChatGPT, demonstrating its robustness and versatility in various applications.
Our dataset and accompanying code are publicly accessible on the repository\footnote{https://github.com/gitdevqiang/SenWave}.
We anticipate that this work will foster further exploration into fine-grained sentiment analysis for complex events within the NLP community, promoting more nuanced understanding and research innovations.
\end{abstract}





\maketitle

\section{Introduction}
The profound global impact of the COVID-19 pandemic has led to significant changes in the lives of individuals worldwide. To mitigate transmission, various measures such as quarantine, curfews, and social distancing have been widely implemented, bringing about notable shifts in work, education, and daily routines.
Analyzing sentiments expressed on social media provides a means to gauge the overall mood of the population. It enables the identification of patterns of fear or anxiety, monitoring public sentiment toward government actions and policies, and detecting emerging concerns or issues \cite{lwin2020global}. This information is invaluable for policymakers, healthcare organizations, and researchers, facilitating informed decision-making, targeted interventions, and effective responses to public concerns \cite{yue2019survey, feng2021integrating, lazzini2022emotions}. Therefore, understanding people's reactions to COVID-19 is crucial for gaining valuable insights into public perceptions and emotional responses to the pandemic.

Despite the abundance of research in natural language processing (NLP) on sentiment analysis during the COVID-19 pandemic \cite{anees2020survey, zhang2018deep, kharde2016sentiment}, significant challenges persist. Two major issues need to be addressed: \textbf{(1) Lack of Comprehensive Annotated Datasets.} Sentiment analysis for COVID-19 requires a substantial volume of tweets with sentiment annotations, covering an extended time window following the outbreak. Despite various datasets, such as the recent one by \cite{xue2020machine} with 1.8 million tweets, no comprehensive dataset for COVID-19 sentiment analysis with large-scale annotations has been established (Table \ref{table:related}). Notably, existing datasets often lack annotations and rely on unsupervised methods based on topic modeling and lexicon features.
\textbf{(2) Lack of Tailored and Fine-Grained Sentiment Labels.} Unlike mainstream sentiment analysis tasks, sentiments surrounding the pandemic are intricate. Existing sentiment analysis typically employs coarse-grained emotion labels like ``positive'', ``neutral'', and ``negative''. However, these labels may not capture the complexity of sentiments during a health crisis. For instance, SemEval-2018, a tweet sentiment dataset with 11 categories, is not well-suited for COVID-19 sentiments. Categories like ``joy'', ``love'', and ``trust'' are underrepresented, and ``official sources'' tweets are misclassified. Additionally, tweets containing jokes or denying conspiracy theories lack appropriate labels. Therefore, incorporating adapted labels such as ``official report'', ``joking'', ``thankful'', and ``denial'' is crucial for effective sentiment analysis in crisis-related tasks.

\begin{table*}[]
\centering
\small
\caption{Summary of recent work on tweets sentimental analysis (None indicates `not used', NA is `not available'). }
\begin{tabular}{p{1.1cm}|p{2.6cm}|p{0.6cm}|p{1.cm}|p{7cm}|p{3.2cm}}
\hline
\multirow{2}{*}{Type}         & \multirow{2}{*}{Related work} & \multicolumn{2}{c|}{\# Tweets}           & \multirow{2}{*}{Sentiment category} & \multirow{2}{*}{Used model/algorithm} \\ \cline{3-4}
&   & \multicolumn{1}{c|}{Labeled} & Unlabeled & &   \\\hline
\multirow{3}{*}{Non-} & Deriu et al. \cite{deriu2016swisscheese} & 18K & 28K & 3 {\small(positive,  neutral, negative) } & CNN+RFC \\ \cline{2-6} 
& Baziotis et al. \cite{baziotis2017datastories}  & 61K & 330M &  3 {\small(positive, neutral,   negative) }&  LSTM+Attention \\ \cline{2-6} 
\multirow{1}{*}{COVID-19}&  \vspace{0.01em} Mohammad et al. {\cite{SemEval2018Task1}}  & \vspace{0.01em}15K & \vspace{0.01em}7,631 &11 {\small(anger, anticipation, disgust, fear, joy, love, optimism, pessimism, sadness, surprise,   trust)}  & Sentence embeddings + lexicons features \\ \hline
& Kabir et al. \cite{kabir2020coronavis} &None & 700GB & 3 {\small(positive, neutral,   negative) }& Topic model (LDA)\\ \cline{2-6} 
 & Xue et al. \cite{xue2020machine} & None & 1.8M  &8 {\small(anger, anticipation, fear, surprise, sadness, joy, disgust,  trust)}  & LDA + NRC Lexicon  \\ \cline{2-6} 
 \multirow{12}{*}{COVID-19}&\vspace{0.01em} Drias et al. \cite{drias2020mining}  & \vspace{0.15em}None &  \vspace{0.01em}65K &10 {\small{(anger, anticipation, disgust, fear, joy, negative ,positive, sadness, surprise, trust)}}  &\vspace{0.01em}Lexicon-based  features \\ \cline{2-6} 
&Kleinberg et al. \cite{kleinberg2020measuring}  & 5K  & None & 8 {\small(anger, anticipation, fear, surprise, sadness, joy, disgust,  trust)} &TF-IDF + POS features  \\ \cline{2-6} 
&Chen et al. \cite{chen2020eyes}  &  2M &  None &2 {\small{(neutral, controversial)}}  & LDA+sentimental dictionary   \\ \cline{2-6} 
 &\vspace{0.01em}Barkur et al. \cite{barkur2020sentiment}  & \vspace{0.01em}None  &\vspace{0.01em} 24K & 10 {\small{(anger, anticipation, disgust, fear, joy, negative ,positive, sadness, surprise, trust)}} & \vspace{0.01em}Lexicon-based features \\ \cline{2-6} 
 
 &Alhajji et al.\cite{alhajji2020sentiment}  & 58K
 & 20K & 2 {\small(positive,  negative) } & Naïve Bayes  \\ \cline{2-6}
 
 &Sri et al. \cite{sri2020mood}  & None & 86K & 6 {\small(anger, disgust, fear, happiness, sadness, surprise) } &Emotion dictionary  \\ \cline{2-6}
 
 &Ziems et al. \cite{ziems2020racism}  &  2.4K &  30K & 3 {\small(hate, counter-hate, neutral) } & Logistic regression classifier  \\ \cline{2-6}
 &Naseem et al. \cite{naseem2021COVIDsenti}  &  90K & None & 3 {\small(positive, neutral, negative) } & BERT\\ \cline{2-6}
 
  & \vspace{0.01em}SenWave (Ours) & \vspace{0.01em}20K & \vspace{0.01em}105M & 10 {\small (optimistic, thankful, empathetic, pessimistic, anxious, sad, annoyed, denial, official report,  joking)} & \vspace{0.01em}BART \\ \hline
\end{tabular}
\label{table:related}
\end{table*}

To address these challenges, we introduce SenWave, a cutting-edge system powered by deep learning, designed specifically for tracking global sentiments during the COVID-19 pandemic. Our team collected 105 million unlabeled tweets related to COVID-19 across five languages: English, Spanish, French, Arabic, and Italian. We annotated 10,000 English tweets and 10,000 Arabic tweets in 10 categories, including \emph{optimistic, thankful, empathetic, pessimistic, anxious, sad, annoyed, denial, official}, and \emph{joking}. The quality evaluation of the annotated data shows the consistent annotations. Additionally, we augmented our dataset by translating the annotated English tweets into Spanish, Italian, and French for broader applicability.
We utilized a transformer-based framework to fine-tune pre-trained language models on the labeled data and revealed intriguing insights into the evolving emotional landscape over time from different aspects of the unlableled data based on the trained model. Our analysis shows a steady increase in optimistic sentiments, aligning with observed trends during the waves of the COVID-19 pandemic. An interesting finding regarding public sentiment toward different parties and policies in the USA demonstrates the value of our dataset for analyzing complex public events.
Furthermore, we leverage ChatGPT to validate the efficacy of our dataset through zero-shot and few-shot multi-label sentiment analysis. Importantly, SenWave offers a unique resource for various sentiment analysis tasks, valuable for the NLP community, especially for complex events requiring fine-grained emotions. 

Our main contributions are summarized below:
\begin{itemize}[leftmargin=*]
\item[a)] We conducted a thorough review of existing sentiment analysis datasets, identifying their limitations, mainly the absence of comprehensive annotated datasets or the deficiency of tailored and fine-grained labels.

\item[b)] We curated the most extensive fine-grained annotated dataset of COVID-19 tweets, featuring 10,000 English and 10,000 Arabic tweets annotated across 10 sentiment categories, as well as 105 million unlabeled tweets. This comprehensive dataset is a valuable resource for exploring the social impact of COVID-19 and facilitating fine-grained analysis tasks within the research community.

 \item[c)] We evaluated the effectiveness of the labeled tweets by fine-tuning transformer-based models compared to several groups of baselines and made predictions on the unlabeled data to analyze the results from different aspects, such as sentiment variation of different languages, countries, topics, and so on. A ChatGPT-based evaluation of our dataset was conducted on zero-shot and few-shot multi-label sentiment analysis to prove the reliability of our dataset.
\end{itemize}

\section{Related work}
\subsection{Non-COVID-19 Tweets based sentiment analysis}
General tweet sentiment analysis often focuses on a few general classes or ordinal sentiment scores ~\cite{srivastava2013quantifying,priyadarshana2015sentiment,balikas2017multitask}. For instance, Sharma et al. classified tweets containing movie reviews into \emph{positive} or \emph{negative} \cite{sharma2020effective}. Baziotis et al. employed LSTM networks with attention mechanisms and pre-trained word embeddings to analyze tweets \cite{baziotis2017datastories}. 
When targeting fine-grained sentiments, the most popular benchmark dataset is SemEval-2018 \cite{SemEval2018Task1}, which includes tweets for gender and race biases prediction\cite{kiritchenko2018examining}. It consists of 7,745 tweets in English, 2,863 in Spanish, and 2,863 in Arabic, labeled across 11 categories.
{\it However, these labels are inadequate for COVID-19 sentiment analysis. For example, there is a scarcity of tweets categorized as ``joy'', and ``love'' while a significant number of tweets from official sources are incorrectly labeled as ``anticipation''.}

\subsection{COVID-19 Tweets based sentiment analysis}
There are numerous public datasets on COVID-19 tweets \cite{kabir2020coronavis,xue2020machine,barkur2020sentiment}. Kabir et al. developed a real-time COVID-19 tweets analyzer to visualize topic modeling results in the USA with three sentiments \cite{kabir2020coronavis}. Kleinberg et al. used linear regression models to predict emotional values based on TF-IDF and part-of-speech (POS) features \cite{kleinberg2020measuring}. Alhajji et al. studied Saudis' attitudes toward COVID-19 preventive measures using Naïve Bayes models to predict three sentiments \cite{alhajji2020sentiment}.
Chen et al. utilized sentiment features and topic modeling to uncover substantial differences in the use of controversial terms in COVID-19 tweets \cite{chen2020eyes}. Ziems et al. applied logistic regression with linguistic features, hashtags, and tweet embeddings to identify anti-Asian hate and counter-hate text \cite{ziems2020racism}.
Barkur et al. used a lexicon-based method to analyze the emotions on the nationwide lockdown of India due to COVID-19~\cite{barkur2020sentiment}.
{\it Despite these advancements, these methods often suffer from coarse-grained sentiments or inappropriate labels and lack data quality evaluation.}

\subsection{Aspect-based Sentiment Analysis}
Aspect-based sentiment analysis (ABSA) focuses on the sentiments of different aspects for the target whose labels are positive, neutral, and negative \cite{hoang2019aspect,zhang2022ssegcn,mahlylyw23}. However, fine-grained sentiment analysis works on a more granular level of labels. For example, in ``The battery life of this phone is amazing, but the camera quality is disappointing. The design is sleek and stylish, though it is a bit expensive.'', ABSA aim to predict the sentiments of ``Battery life'', ``Camera quality'', ``Design'', and ``Price'', corresponding to positive, negative, positive, and negative, respectively. Hoang et al. proposed to fine-tune BERT to a sentence pair classification model for ABSA \cite{hoang2019aspect}. 
Zhang proposed a SSEGCN architecture which integrated semantic information along with the syntactic structure for ABSA task by combining attention scores with syntactic masks \cite{zhang2022ssegcn}.

\subsection{Comparison with Existing Works}
The literature on sentiment analysis can be categorized into coarse-grained (e.g., positive, negative, neutral) and fine-grained emotions (e.g., optimistic, pessimistic, anxious, fear, joy, happiness) domains. Our work presents a more nuanced approach, addressing the following challenges: 1) Complexity of COVID-19 Events: The COVID-19 pandemic is a multifaceted public health event, demanding a nuanced understanding of public sentiment. Labels like positive, neutral, and negative are insufficient for capturing the complexity. 2) Suitability for COVID-19 Context: Existing labels such as joy, love, trust, amusement, pride, joy, and love may not be suitable for COVID-19-related discussions. 3) Specialized Labels: Labels like hate and counter-hate are specialized for anti-Asian hate and counter-hate detection during the COVID-19 crisis. Similarly, COVID-19 events exhibit unique characteristics, including a significant number of official reports released by governments, as well as ironic or sarcastic sentiments.

\section{Dataset Construction}
\label{sec:dataset}
\subsection{Data Collection}
We employed Twint\footnote{https://github.com/twintproject/twint}, an open-source Twitter crawler, to collect tweets. Twint offers flexibility by allowing users to specify parameters, including tweet language and time. We focused on five languages: English, Spanish, French, Arabic, and Italian. The search terms used across these languages included ``COVID-19'', ``COVID19'', ``coronavirus'', ``COVID'', ``corona'', and corresponding Arabic terms. Retweets were included in our dataset as they often contain additional user-generated content, such as comments or opinions, which can be valuable for sentiment analysis.
To efficiently gather the data, we deployed 12 instances of Twint on a workstation equipped with 24 cores, downloading daily updates from March 1 to May 15, 2020. The data was saved as JSON documents for subsequent pre-processing. More data will be released regularly for updates and maintenance.

\begin{table*}[h]
    \centering
    \small
    \caption{The label distributions of the annotated English, and Arabic datasets (\%).} 
    \begin{tabular}{c|c|c|c|c|c|c|c|c|c|c}\hline
        & Opti. &  Than. & Empa. & Pess. & Anxi. & Sad &Anno. & Deni. & Offi. &Joki.\\\hline
       English &23.73& 4.98& 3.89& 13.25& 16.95& 21.33& 34.92& 6.31& 12.07& 44.76\\\hline
       Arabic &11.27& 3.33& 6.49& 4.65& 7.53& 10.80& 17.17& 2.10& 34.52& 14.18\\\hline
    \end{tabular}
    \label{tab:labdist}
\end{table*} 

\subsection{Data Annotation}
After collecting unlabeled tweets, we performed sentiment annotation on a randomly selected subset of 10,000 English and 10,000 Arabic tweets.
\\
\textbf{Sentiment Categories Determination.}
We enlisted the expertise of four domain experts with a rich background in public health and epidemiology. The experts carefully reviewed a subset of the collected tweets, drawing inspiration from SemEval-2018, to determine the ten sentiment categories that encompass the complex range of emotions observed during the pandemic. These labels include: \textit{optimistic} (representing hopeful, proud, and trusting emotions), \textit{thankful} (expressing gratitude for efforts to combat the virus), \textit{empathetic} (including prayers and compassionate sentiments), \textit{pessimistic} (reflecting a sense of hopelessness), \textit{anxious} (conveying fear and apprehension), \textit{sad}, \textit{annoyed} (expressing anger or frustration), \textit{denial} (towards conspiracy theories), \textit{official report} (the release of factual information by governments or official organizations, such as confirmed cases, deaths, vaccine doses administered, and epidemic prevention policies), and \textit{joking} (irony or humor).

The inclusion of the ``official report'' category stems from the fact that governments and organizations like the WHO often release factual information about COVID-19, including confirmed cases, deaths, vaccine doses administered, and epidemic prevention policies. These tweets do not fit neatly into ``positive'', ``negative'', or ``neutral'' labels; rather, they represent objective reporting of facts.
\\
\textbf{Annotation Process.}
Our data was labeled by Lucidya\footnote{https://lucidya.com/}, an AI-based company with extensive experience in organizing data annotation projects. Each tweet was independently labeled by three annotators, who were compensated at a rate of 0.6 USD per tweet. To ensure reliable annotations, we recruited 52 experienced annotators who were native or fluent speakers. They were trained with example tweets and suggested categories to guide the annotation process, using an annotation guideline notebook (available on our anonymous GitHub \footnote{https://anonymous.4open.science/r/SenWave-36BE/README.md}).
We allowed multi-label annotation to capture the nuanced and complex emotions experienced during the pandemic. The final labels for tweets were determined by majority voting when the annotations overlapped. Otherwise, the tweets were marked and re-annotated to achieve consistent results. Ultimately, we obtained 10,000 annotated English tweets and 10,000 annotated Arabic tweets, each labeled across ten categories.
\\
\textbf{Annotation Quality Evaluation.}
Following the methodology of \cite{SemEval2018Task1}, we used Average Inter-rater Agreement (IRA) and Cohen's Kappa Coefficient to assess the quality and agreement of the sentiment annotations. IRA measures the average percentage of times each pair of annotators agrees, while Cohen's kappa coefficient assesses inter-rater reliability for qualitative items. Inter-annotator agreement scores were calculated by computing the scores for each pair of annotators three times and then averaging these scores. The $\iota$ values for English and Arabic annotations reached 0.904 and 0.931, while the Kappa coefficients $\kappa$ were 0.381 and 0.549, respectively. The high IRA values indicate a substantial level of agreement among the annotators, and the good Cohen's kappa coefficients demonstrate fair and moderate agreement for our labeled data.

\subsection{Data Augmentation}
Considering the advancements in translation tools, we translated the labeled English tweets into Spanish, French, and Italian using Google Translate to augment our dataset. Data augmentation offers several benefits. (1) \textit{Increased Diversity}: It enhances the dataset's diversity by recognizing sentiment expressions in different linguistic and cultural contexts. (2) \textit{Scalability}: It provides a scalable way to create a larger training dataset without the need for manual labeling. (3) \textit{Cost-Effectiveness}: It serves as a cost-effective alternative to leverage existing labeled data for multiple languages.
To evaluate the quality of translation, we calculated the BLEU score by comparing A and A', where A' is translated back from A(En) to B(Es) to A'(En), using English and Spanish as examples. The BLEU score was 0.33 (noting that the SOTA machine translation model has a BLEU4 score of 0.39 using a tied transformer), verifying the good quality.

\subsection{Data Overview}
\label{sec:data}
In this section, we provide an overview of the basic information regarding both the labeled and unlabeled tweets in our dataset.
\subsubsection{Annotated Tweets}
\begin{figure}[ht!]
\centering
\includegraphics[width=.44\textwidth]{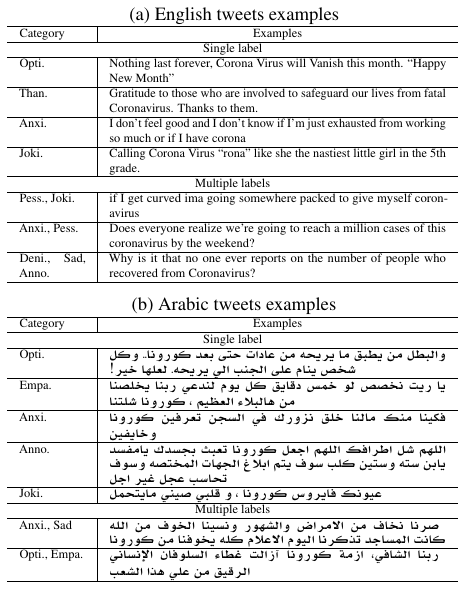}
\caption{Random Examples of Labeled Tweets}
\label{fig:exam}
\end{figure}

\begin{figure}[h]
\centering 
\subfigure[English tweets]{
\includegraphics[width=0.23\textwidth]{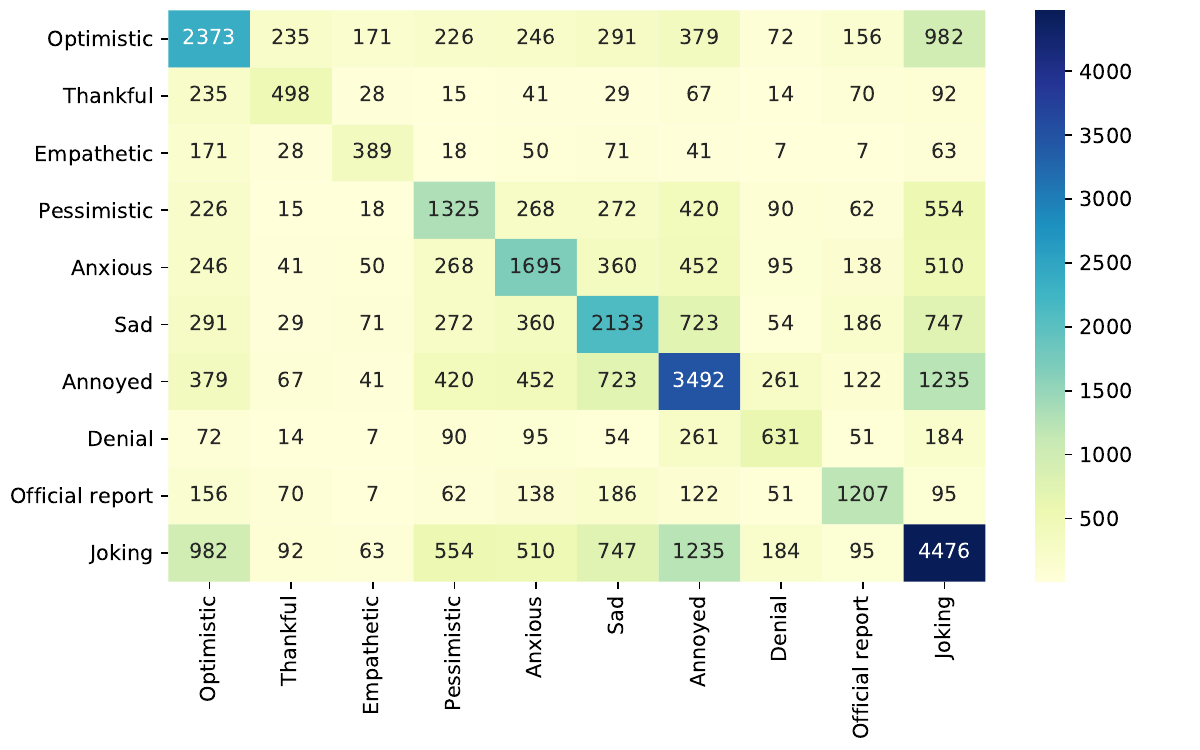}}
\subfigure[Arabic tweets]{
\includegraphics[width=0.23\textwidth]{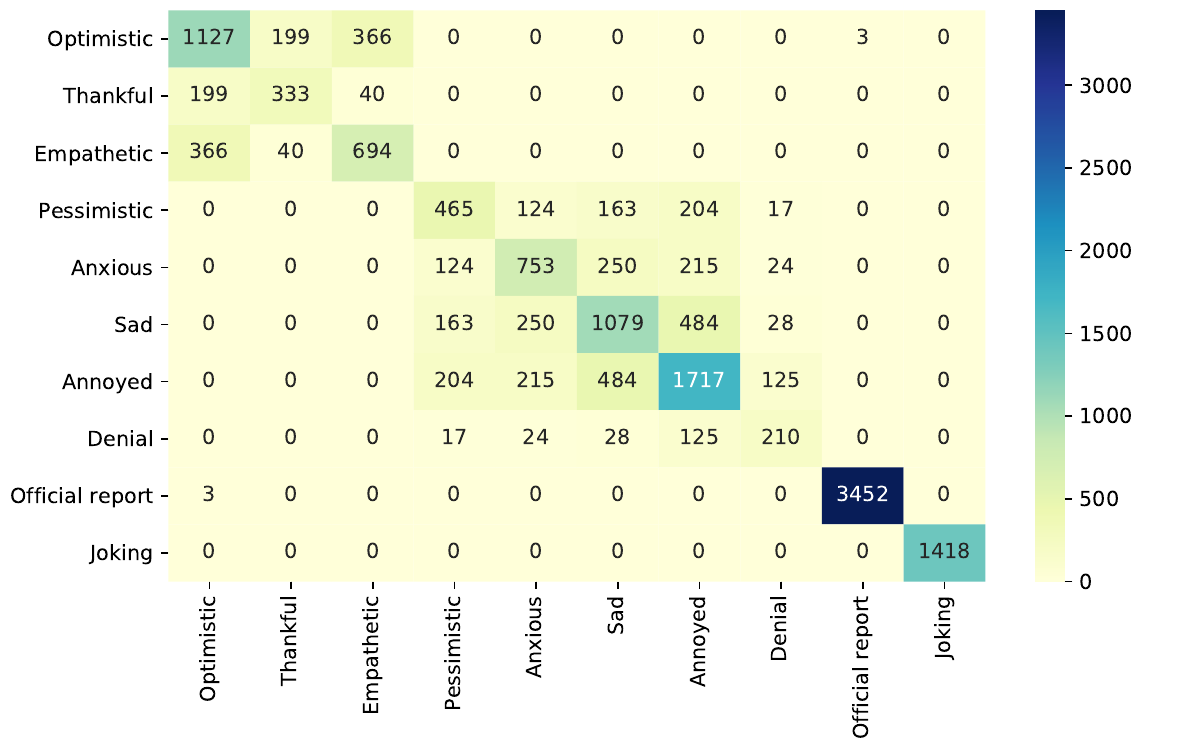}}
\vspace{-0.1in}
\caption{Heatmaps of labels co-occurrence for English and Arabic tweets.
}
\label{fig:hm}
\end{figure}

\leavevmode\newline
\textbf{Data Distribution.} 
The distribution of labels for each sentiment category in the annotated English and Arabic tweets is detailed in Table \ref{tab:labdist}. In the English dataset, emotions such as \emph{joking} and \emph{annoyed} dominate, reflecting the harsh realities of COVID-19, including fatalities, high unemployment rates, and other challenges. Surprisingly, the \emph{optimistic} emotion represents the third-largest category, suggesting that people also maintain confidence and hope in overcoming the virus and envisioning a positive future.
In the Arabic dataset, the ``official report'' label stands out significantly, attributed to numerous announcements and decisions made by Arabic governments in response to the outbreak. Differences in label distribution between English and Arabic tweets may be influenced by distinct cultural backgrounds and religions. The prevalence of the {\it joking} label is higher in English tweets compared to Arabic, while the {\it empathetic} label exhibits the opposite trend. This pattern is reflected in predictions on unlabeled data with our classification model.
Additionally, the percentages of all labels do not sum up to 100\%, due to the multi-label annotation in our dataset.
\\
\textbf{Data Examples.}
Figure \ref{fig:exam} provides examples of annotated English and Arabic tweets. A preliminary analysis of the labeled data reveals that in English tweets, over 70\% feature multiple labels, whereas in Arabic tweets, approximately 20\% exhibit the same characteristic. Therefore, the sentiment analysis task on English tweets is more challenging than on Arabic tweets, as demonstrated in the experimental section.
\\
\textbf{Label Co-occurrence Relations of English and Arabic Data.}
We utilized label co-occurrence heatmaps to illustrate the interrelationships between sentiment labels in both the English and Arabic datasets. In Fig. \ref{fig:hm} (a), the complexity of label co-occurrence in the English dataset was evident, underscoring the intricacy of multi-label classification challenges. On the other hand, Fig. \ref{fig:hm} (b) revealed that the sentiment ``official report'' predominates in the Arabic dataset, reflecting the substantial influence of decisions made by the Saudi government. This disparity in label co-occurrence patterns highlights the nuanced nature of sentiment expression across different languages and cultural contexts.

\subsubsection{Unlabeled Tweets}
\begin{figure}[ht!]
\centering
\includegraphics[width=.45\textwidth]{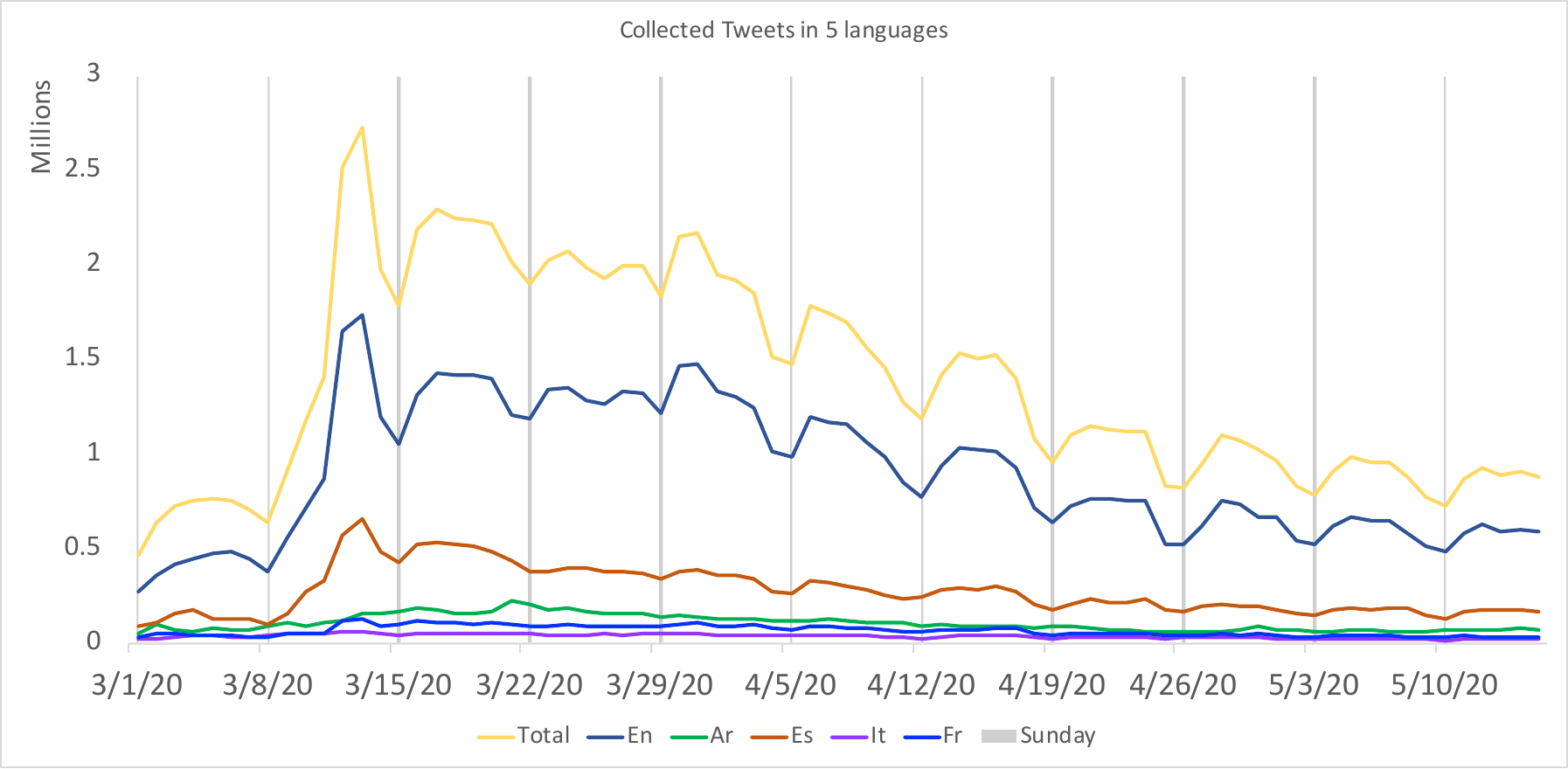}
\caption{The absolute daily volume of COVID-19 Tweets collected in 5 languages, English (En), Spanish (Es), Arabic (Ar), French (Fr), and Italian (It). The vertical lines show Sundays, for guidance.}
\label{fig:vol}
\end{figure}

\leavevmode\newline
\textbf{Data Volume.}
We collected 105 million unlabeled tweets related to COVID-19, spanning from March 1 to May 15, 2020, covering the first wave of the pandemic. The tweets were gathered in five languages: English, Spanish, French, Arabic, and Italian. The daily volume of collected tweets for each language is depicted in Fig.~\ref{fig:vol}.
English tweets dominated with the largest number, followed by Spanish tweets, and then Arabic tweets, reaching their daily maximum on March 13 or March 21. These peaks coincided with significant events such as the US President declaring a national emergency, the Spanish Prime Minister declaring a state of emergency, and Saudi Arabia suspending public travel.
\\
\textbf{Data Glance.} 
The trend of people's attention shows an initial increase to a peak point, followed by a gradual decline over time. This pattern is consistent across different languages, indicating a similar response to the pandemic among speakers of various languages. These characteristics underscore the reliability of our collected data.
Interestingly, the number of tweets shows a downward trend on Sundays. A possible reason is that Sundays are typically the weekend in many cultures, and people may be engaged in activities that do not involve as much social media usage, such as enjoying time with family and participating in leisure activities.

\section{Sentiment Classification Models}
\label{sec:model}
\subsection{Data Preprocessing and Anonymization}
To prepare the raw tweets for sentiment analysis, we initiated the process with several preprocessing steps. Initially, we eliminated URLs as they do not contribute significantly to the sentiment analysis. Emojis and emoticons, such as $\ddot\smile$, were also removed, despite their expressive nature, as our focus was on analyzing textual data. Subsequently, we filtered out noisy symbols and texts that lack meaningful semantics, including the retweet symbol "RT" and special characters like line breaks, tabs, and redundant blank spaces. Unlike some prior methods, we retained hashtags in tweets, as they often encapsulate the primary theme or topic of the tweet, facilitating a better understanding of the subject matter. Additionally, we performed word tokenization, stemming, and tagging using NLTK\footnote{https://www.nltk.org/} for English, Spanish, French, and Italian, and Pyarabic\footnote{https://pypi.org/project/PyArabic/} for Arabic. User-relevant information, such as usernames, gender, and strings starting with the character @, was also removed to protect user privacy.

\subsection{Multi-label Sentiment Classifiers}
Our multi-label sentiment classifier, rooted in the success of the Transformer architecture across various NLP tasks, was crafted by fine-tuning language models with a customized classifier featuring two MLP layers. Specifically, we leveraged BART \cite{lewis2019bart} for English, AraBERT \cite{antoun2020arabert} for Arabic, and BERT \cite{devlin2018bert} for Spanish, French, and Italian. To evaluate the effectiveness of our approach, we compared it with several baselines, including Fasttext, CNN, LSTM, LSTM-CNN, CNN-LSTM, BERT, BERTweet, and XLNet, all using the same classifier layers as ours. Non-Transformer-based methods employed 300-dimensional Glove embeddings for word representations. We trained the models with binary cross-entropy loss.

\subsection{Experimental Settings and Metrics}
The experiments were conducted on a workstation equipped with one GeForce GTX 1080 Ti. The training setup included a batch size of 16, a learning rate of $4e-5$, and the models were trained over 20 epochs. The Adam optimizer was employed, and a fixed random seed of 42 was used for consistency.
To assess performance, metrics such as multi-label accuracy, F1-macro, F1-micro, ranking average precision score (LRAP), and Hamming loss were employed. The evaluation was carried out using 5-fold cross-validation to ensure a robust assessment of model performance.
%
%

\begin{table}[t]
    \centering
    \small
    \caption{\small{(a) Overall validation on the SenWave dataset.} }
    \vspace{-0.1in}
    \begin{tabular}{p{0.2cm}|p{1.3cm}|p{1.3cm}|p{1.3cm}|p{1.3cm}|p{1.3cm}}\hline
          &Accuracy& F1-Macro & F1-Micro & LRAP & Hamm.Loss\\\hline
       En & 0.498$\pm$0.008 &0.535$\pm$0.012& 0.580$\pm$0.008 & 0.548$\pm$0.007& 0.156$\pm$0.004 \\\hline
       Ar & 0.591$\pm$0.010 & 0.488$\pm$0.016&0.614$\pm$0.008&0.635$\pm$0.009&0.083$\pm$0.002\\\hline 
       Sp &0.428$\pm$0.004&0.434$\pm$0.010 &0.511$\pm$0.003 &0.493$\pm$0.002&0.177$\pm$0.001 \\\hline
       Fr &0.430$\pm$0.010&0.432$\pm$0.010&0.509$\pm$0.010&0.496$\pm$0.009&0.176$\pm$0.004 \\\hline
       It & 0.437$\pm$0.006 &0.442$\pm$0.010 &0.517$\pm$0.005 &0.503$\pm$0.005&0.172$\pm$0.002 \\
       \hline 
    \end{tabular}    
    \label{tab:perf}
    \quad
    \vspace{0.1in}
    \\ {\small \textbf{(b) Accuracy of each category on the SenWave dataset.}}
    \small
    \begin{tabular}{p{0.55cm}|p{1.2cm}|p{1.2cm}|p{1.2cm}|p{1.2cm}|p{1.2cm}}
    \hline
        ~ & \centering{En} & \centering{Ar} & \centering{Sp} & \centering{Fr} & It  \\\hline
        Opti. & 0.441$\pm$0.012 & 0.418$\pm$0.025& 0.329$\pm$0.011 & 0.319$\pm$0.013&  0.333$\pm$0.007 \\ \hline
        Than. & 0.290$\pm$0.020 & 0.425$\pm$0.038 & 0.183$\pm$0.028& 0.167$\pm$0.021 & 0.166$\pm$0.025  \\ \hline
        Empa. & 0.438$\pm$0.018 & 0.459$\pm$0.042 & 0.243$\pm$0.032& 0.278$\pm$0.024 & 0.292$\pm$0.056  \\ \hline
        Pess. & 0.194$\pm$0.022 & 0.116$\pm$0.039 & 0.101$\pm$0.024& 0.094$\pm$0.016 &  0.101$\pm$0.010 \\ \hline
        Anxi. & 0.309$\pm$0.021 & 0.222$\pm$0.033 & 0.219$\pm$0.015&  0.216$\pm$0.025  & 0.229$\pm$0.008 \\ \hline
        Sad & 0.309$\pm$0.018 & 0.254$\pm$0.020 & 0.250$\pm$0.010& 0.241$\pm$0.014 &  0.233$\pm$0.022 \\ \hline
        Anno. & 0.514$\pm$0.016 & 0.389$\pm$0.032 & 0.429$\pm$0.010& 0.428$\pm$0.023 & 0.430$\pm$0.014  \\ \hline
        Deni. & 0.249$\pm$0.023 & 0.116$\pm$0.051 & 0.150$\pm$0.014& 0.141$\pm$0.008 & 0.166$\pm$0.023  \\ \hline
        Offi. & 0.619$\pm$0.019 & 0.872$\pm$0.017 & 0.566$\pm$0.017& 0.569$\pm$0.025&  0.576$\pm$0.022 \\ \hline
        Joki. & 0.559$\pm$0.022 & 0.358$\pm$0.027 & 0.514$\pm$0.019& 0.516$\pm$0.012 & 0.522$\pm$0.023  \\ \hline
    \end{tabular}
\end{table}
\section{Results and Analysis}	
\subsection{Multi-label Classifier Results}
The evaluation results of the sentiment classifiers are summarized in Table \ref{tab:perf} (a). We found that the performance on the Arabic data outperformed that on the English data, which can be attributed to a higher rate of multiple labels in English tweets compared to Arabic tweets. This suggests that classifying English tweets is relatively challenging.
The accuracy of Spanish, French, and Italian tweets was lower than that of the original data. This can be explained by the use of different pre-trained language models: BART and AraBERT perform better than the generally used BERT for Spanish, French, and Italian under the same conditions \cite{yang2019xlnet,antoun2020arabert}. F1 values around 0.5 were influenced by the issue of class imbalance.
The accuracy of each sentiment category, as shown in Table \ref{tab:perf} (b), revealed that \textit{official report}, \textit{joking}, \textit{optimistic}, and \textit{annoyed} can be predicted with higher accuracy. On the other hand, \textit{pessimistic} and \textit{thankful} were more challenging to predict. 
In the comparison with baselines in Table \ref{tab:baseline}, BART performed the best among all models, followed by BERTTweet, XLNet, and BERT, all of which belong to the Transformer group. FastText and CNN-LSTM exhibited similar performance, where FastText showed better out-of-vocabulary (OOV) capabilities compared to GloVe, and CNN captures local semantics better than LSTM. These results underscore the effectiveness of Transformer-based models.
\subsection{Dataset Reliability Evaluation}
To validate the usability of SenWave, we employed GPT-3.5 for multi-label text classification on English data. We conducted tests in both zero-shot and few-shot learning scenarios. As shown in Table \ref{tab:gpt}, the performance of few-shot text classification outperformed that of zero-shot classification across all metrics. This indicates two key findings: 1) Our dataset is effective for multi-label text classification, capable of capturing complex and nuanced sentiments. 2) It can be employed for low-resource tasks involving complex sentiments, highlighting its versatility and utility in diverse research contexts.
The prompts used for the fine-grained sentiment analysis are provided in Appendix B.

\begin{table}[]
    \caption{\small{ Comparison of all models on the SenWave dataset.}}
    \small
    \vspace{-0.1in}
    \begin{tabular}{c|c|c|c|c|c}
    \hline
        Models & Accuracy & F1-Macro & F1-Micro & LRAP & Hamm.Loss \\ \hline
        Fastext & 0.371 & 0.269 &0.453 & 0.469& 0.162 \\ \hline
        CNN &0.389 & 0.387& 0.482& 0.470&0.178\\ \hline
        LSTM & 0.328& 0.369& 0.419& 0.399&0.231\\ \hline
        LSTM-CNN & 0.312& 0.380& 0.413& 0.368&0.264\\ \hline
        CNN-LSTM &0.361 &0.411 &0.453 & 0.430&0.207\\ \hline
        BERT &0.479 &0.506 &0.571 & 0.530&0.159\\ \hline
        BERTTweet & 0.498&0.535 & 0.585& 0.542&0.159\\ \hline
        XLNet &0.495 &0.517 & 0.573 & 0.535&0.153\\ \hline
        BART & 0.498&0.535 & 0.580&0.548 &0.156\\ \hline
    \end{tabular}
    \label{tab:baseline}
\end{table}

\begin{table}[]
    \centering
    \small
    \caption{Zero-shot and Few-shot Text Classification with ChatGPT on English Dataset}
    \vspace{-0.1in}
     \begin{tabular}
     {c|c|c|c|c|c}\hline
        &Accuracy& F1-Macro & F1-Micro & LRAP & Hamm.Loss\\\hline
        Zero-shot &0.137&0.238 & 0.275&0.377 &0.212\\ \hline
        Few-shot &0.190 &0.309 & 0.386 & 0.430&0.200\\ \hline
    \end{tabular}
    \label{tab:gpt}
\end{table}

\subsection{Sentiment Variation of Unlabeled Tweets}
In this section, we explore the variation of sentiments in different contexts, including {\bf different languages}, {\bf different countries}, {\bf different topics}. We also conduct the analysis on {\bf the emotion of Joking} and {\bf public attitudes towards political parties}.
\paragraph{1)  Sentiment Variation in Different Languages Over Days.}
\begin{figure}[h]
    \includegraphics[width=0.48\textwidth]{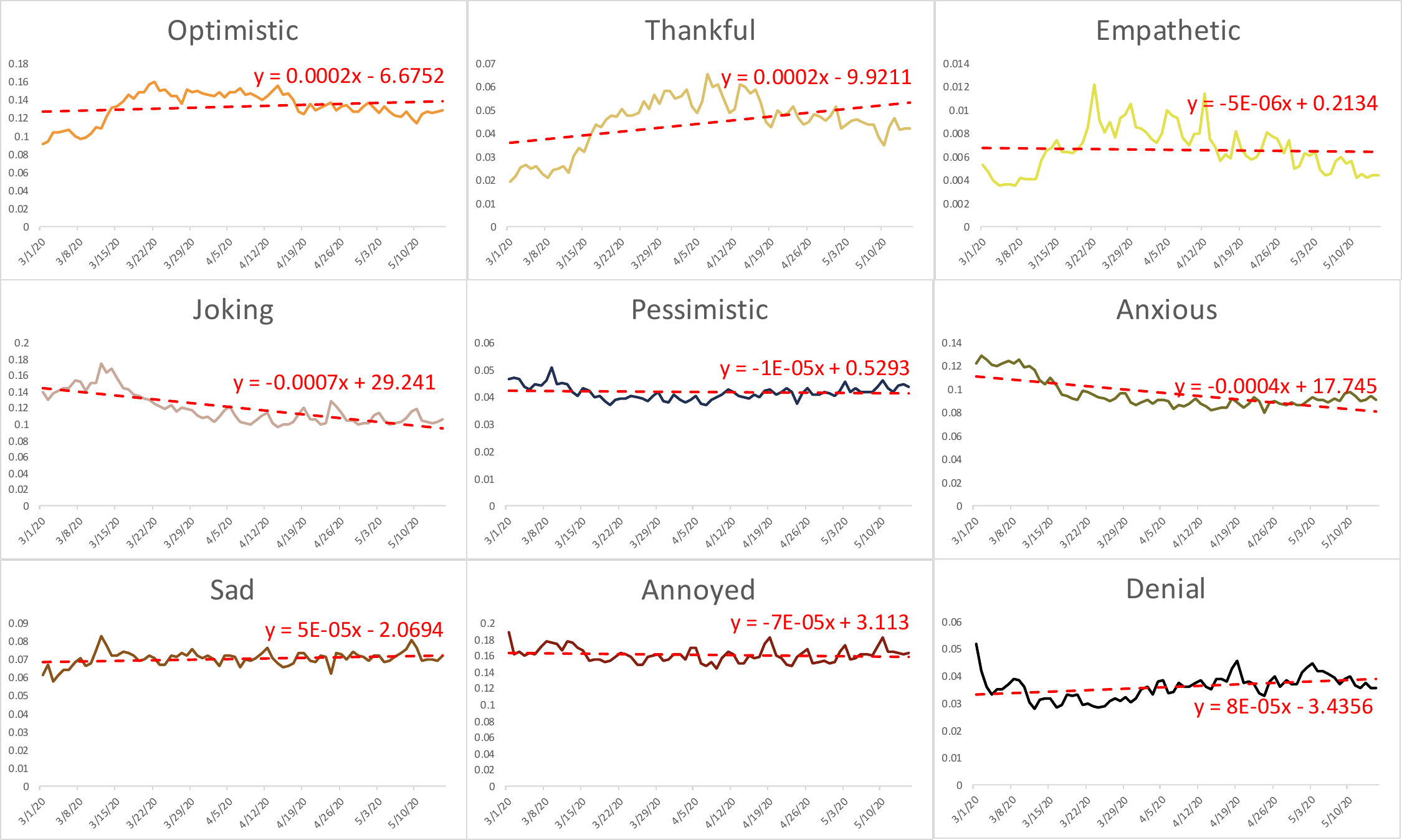}
    \vspace{-0.1in}
\caption{Sentiment variation of English tweets over time. The linear regression line of each emotion curve shows the trend of the emotion variation.}
\label{fig:lan0}
\end{figure}
We illustrated the sentiment variation of English tweets in Fig.~\ref{fig:lan0}. All positive emotions exhibited a similar trend of initially rising and then declining. This suggests that people initially felt positive due to various decisions made to combat the virus in mid-March. However, these emotions declined in late April when more people were infected. Among negative emotions, \emph{anxious} and \emph{joking} decreased over time. The decrease in \emph{anxious} may be attributed to an increase in medical supplies, while the persistently high levels of \emph{sad} and \emph{annoyed} could be linked to the rising unemployment rate and death toll. Results for other languages are provided in Appendix C.1.

\begin{figure}[htb]
    \centering
    \includegraphics[width=0.48\textwidth]{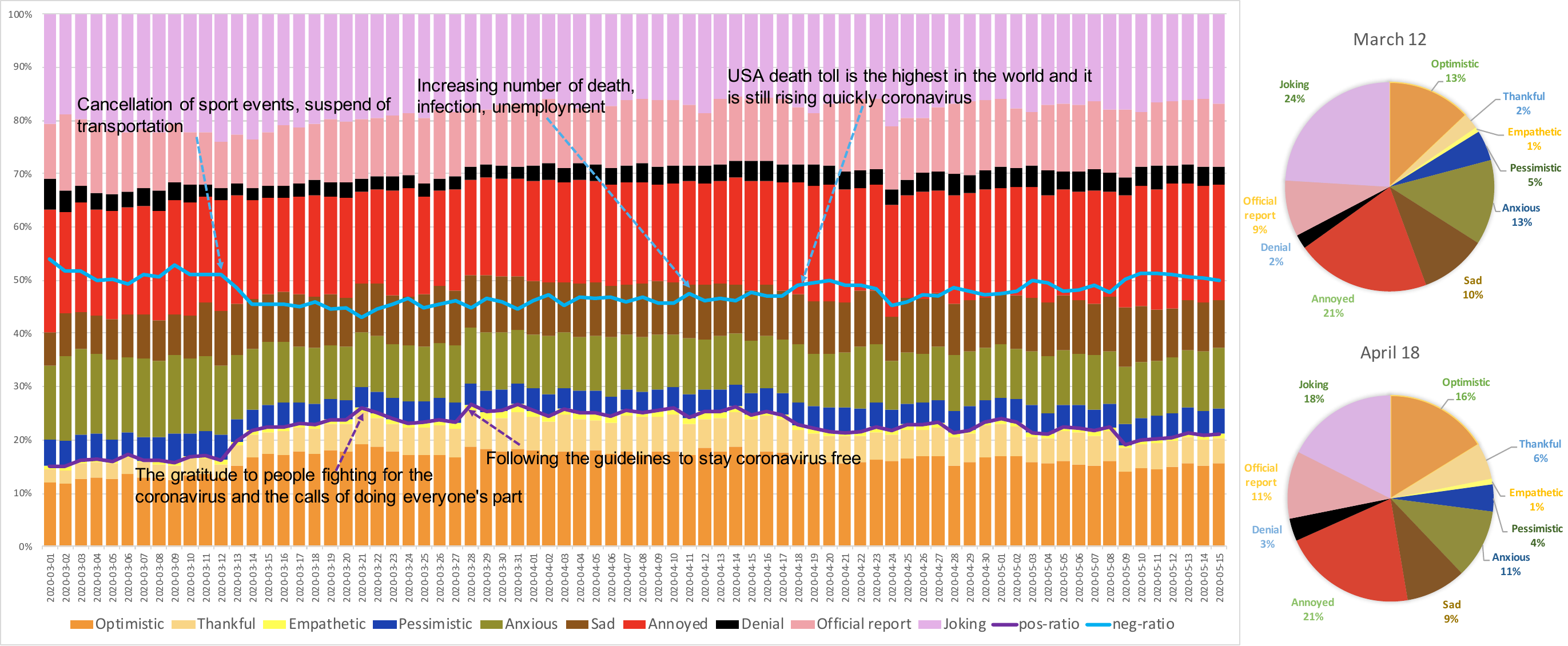}
    \vspace{-0.1in}
    \captionof{figure}{Sentiment variation in USA over time. Each bar shows the distribution of sentiments on one day (Better zoom in the spikes).}
    \label{fig:areas0}
\end{figure}

\paragraph{2) Sentiments Variation of Different Countries Over Days.}
We chose the USA as an example to illustrate how sentiments vary over days in Fig.~\ref{fig:areas0}. The blue and purple curves represent positive (sum of \emph{optimistic, thankful, empathetic} in yellow at different intensities) and negative (sum of \emph{pessimistic, anxious, sad, annoyed, denial} in blue at different intensities) sentiments, respectively. We observed that the proportion of negative emotions was consistently higher than that of positive emotions. On March 12, people expressed \emph{annoyed} and \emph{anxious} sentiments (see the pie charts) as normal life was affected by the coronavirus, including the cancellation of sports events and suspension of transportation. On March 21, positive emotions slightly increased as people expressed gratitude for the efforts of healthcare workers. However, negative emotions rose again due to increasing rates of death, infection, and unemployment on April 11. Results for other countries are provided in Appendix C.2.

\begin{figure}[htb]
    \centering
    \includegraphics[width=0.48\textwidth]{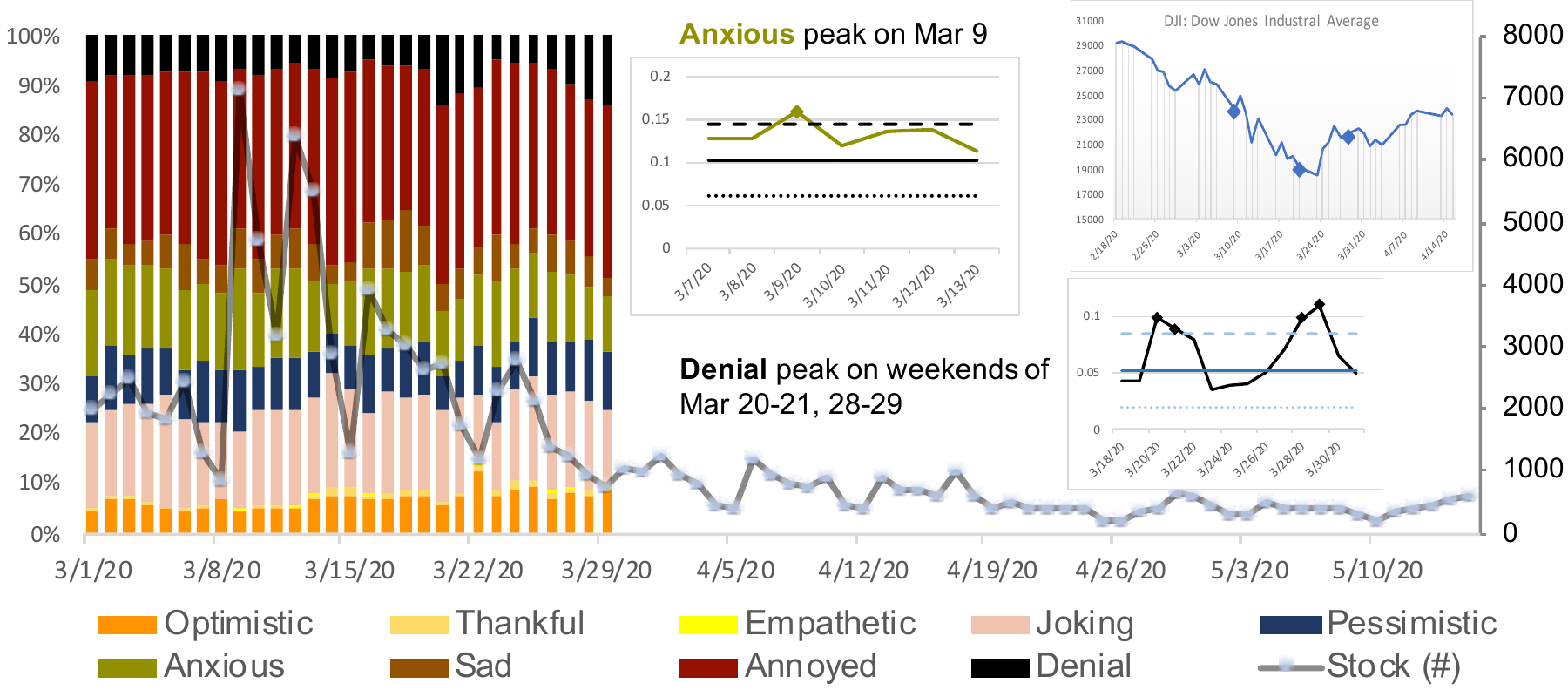}
    \vspace{-0.1in}
    \captionof{figure}{Sentiments variation on the stock market. We show the sentiment results when the topics were intensively discussed (around the peak of the volume curve in the background.}
    \label{fig:topics0}
\end{figure}

\begin{figure}[h]
\centering 
\subfigure[]{
\includegraphics[width=0.23\textwidth]{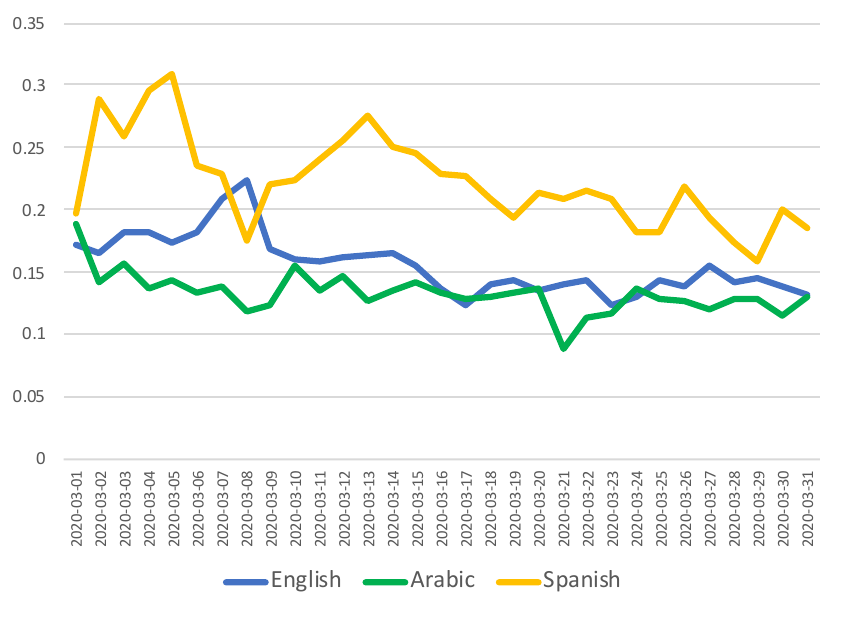}}
\subfigure[]{
\includegraphics[width=0.23\textwidth]{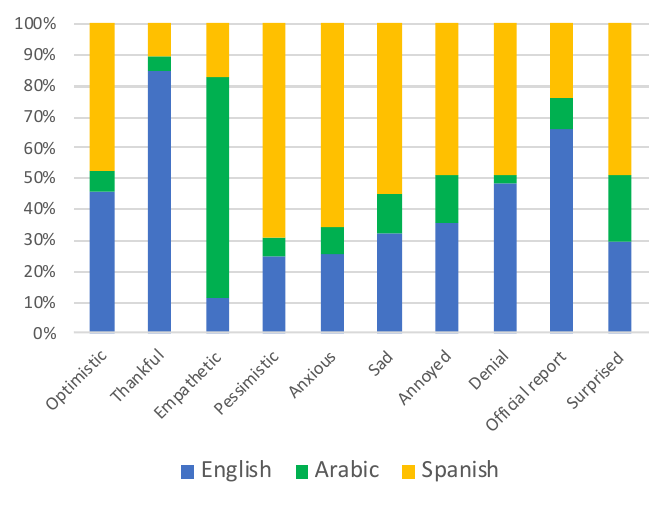}}
\subfigure[]{
\includegraphics[width=0.3\textwidth]{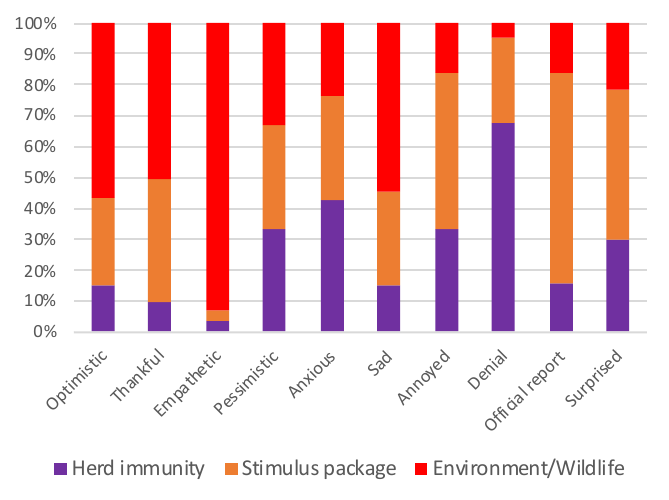}} 
\vspace{-0.1in}
\caption{Analysis of the category \emph{joking}. (a) The portion of \emph{joking} overtime in 3 languages. (b) and (c) show the co-occurrence of joking and other labels in 3 languages and 3 events, respectively.
}
\label{fig:joking}
\end{figure}

\paragraph{3) Sentiments Variation of Topics Over Days.}
We analyzed the sentiment regarding the topic \emph{stock market} in Fig.~\ref{fig:topics0}. The stock market collapsed on March 9 when the peak of the discussion was reached. \emph{Anxious} reached a high value, surpassing the mean+2std (above the black dashed line, where the black line represents the mean, and the dotted line is the mean-2std). On March 12, the DJI (Dow Jones Index) experienced its worst day since 1987, plunging about 10\% (triggering the second circuit breaker). On the weekends of March 20-21 and March 28-29, the spikes of \emph{denial} were higher than the blue dashed line, reflecting the collapse of the stock market. Results for more topics, such as herd immunity and economic stimulus, are discussed in Appendix C.3.

\begin{figure}[h]
\centering 
\subfigure[]{
\includegraphics[width=0.22\textwidth]{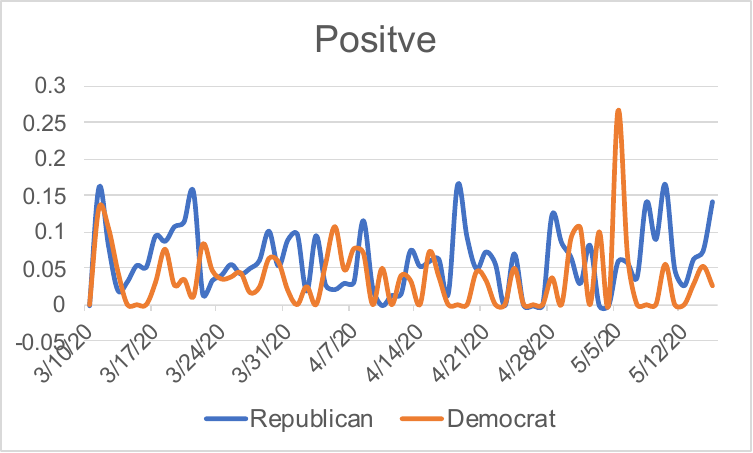}}
\subfigure[]{
\includegraphics[width=0.22\textwidth]{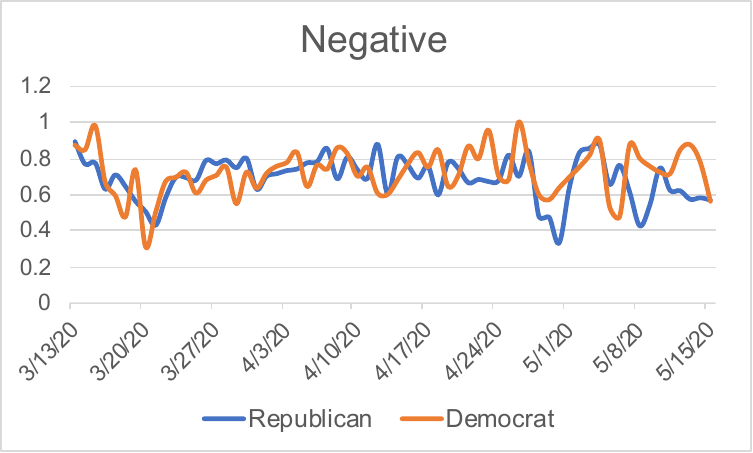}}
\subfigure[]{
\includegraphics[width=0.22\textwidth]{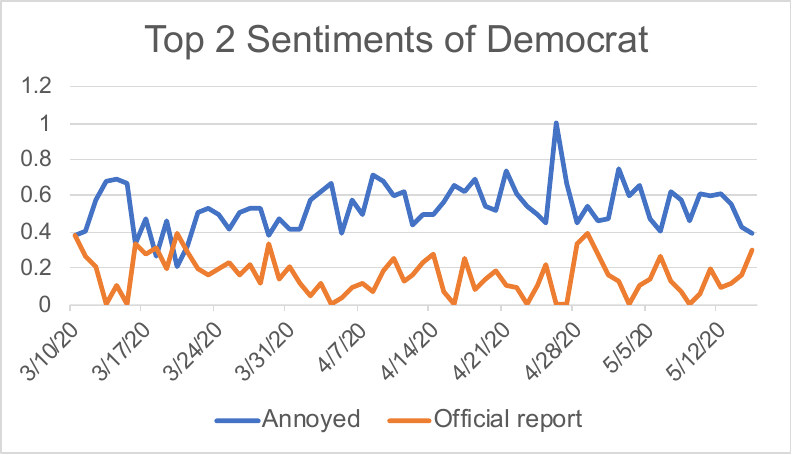}} 
\subfigure[]{
\includegraphics[width=0.22\textwidth]{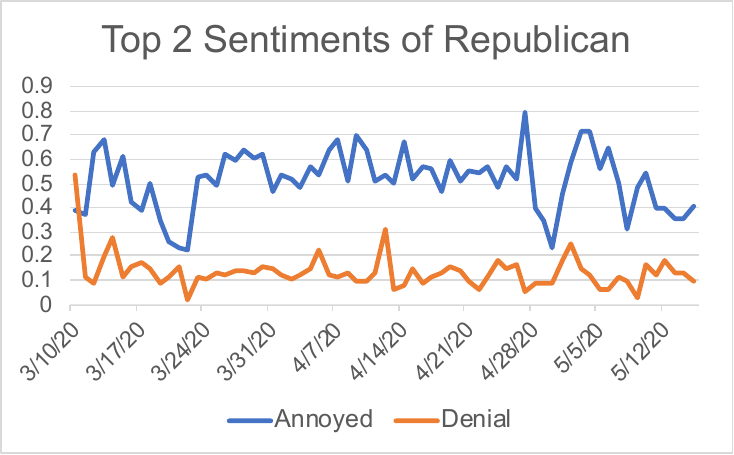}} 
\vspace{-0.1in}
\caption{Analysis of public’s attitude towards two political parties. (a) and (b) are the trend of positive and negative sentiment. (c) and (d) show the top two sentiments over time for political parties.
}
\label{fig:parties}
\end{figure}

\begin{figure*}[]
\centering  
\subfigure[Optimistic]{
\label{enwcop}
\includegraphics[width=0.18\textwidth]{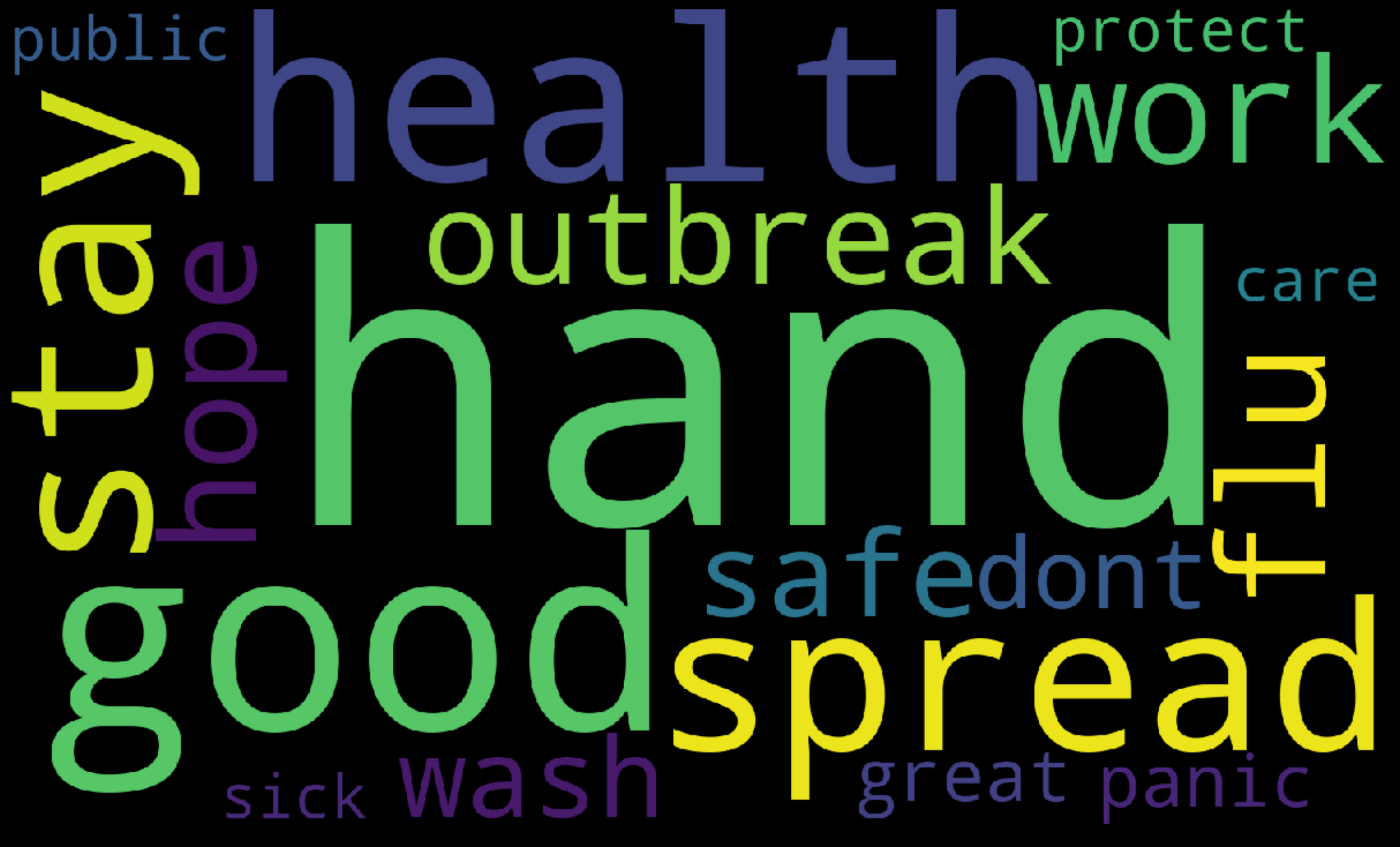}}
\subfigure[Thankful]{
\label{enwcth}
\includegraphics[width=0.18\textwidth]{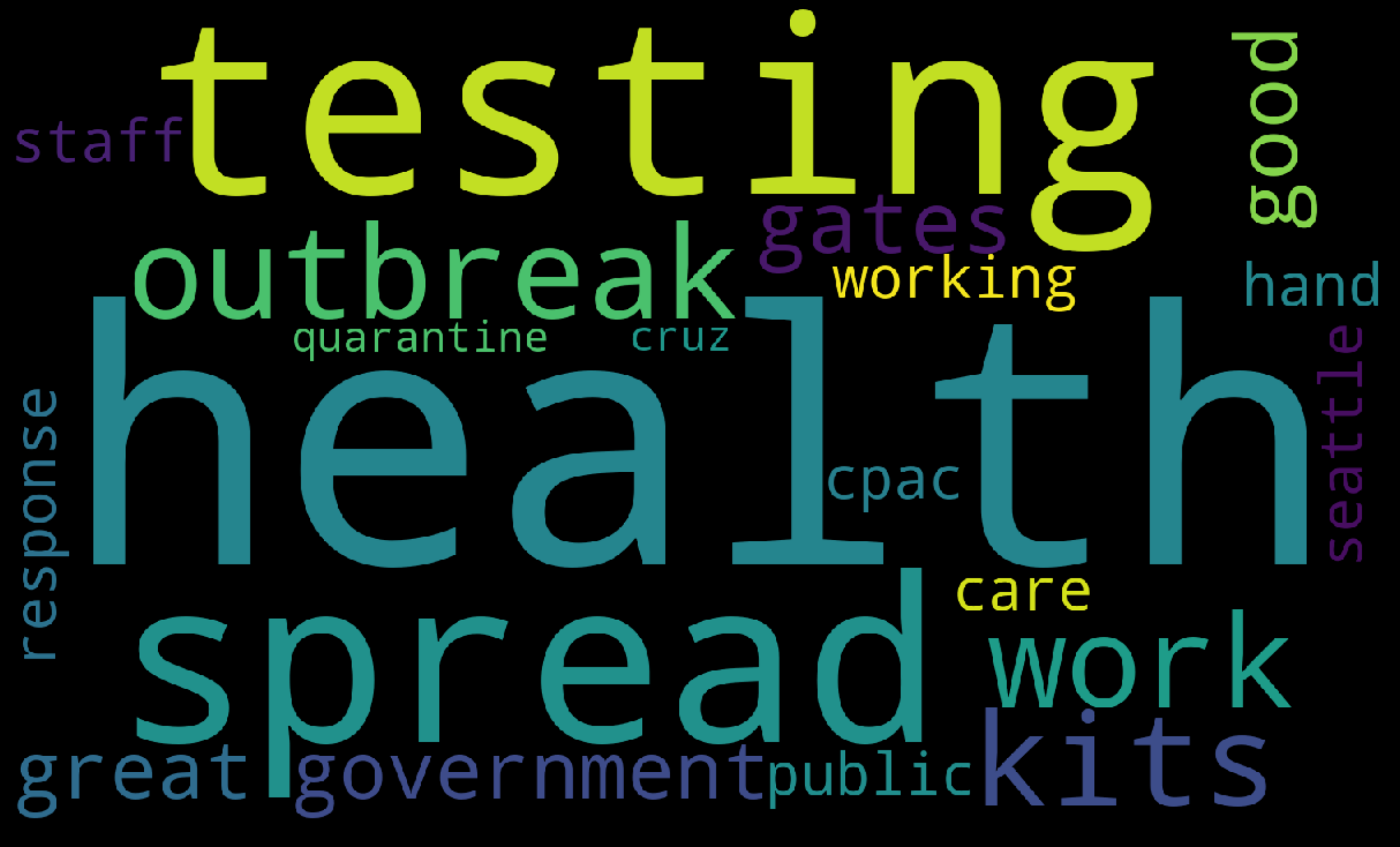}}
\subfigure[Empathetic]{
\label{enwcem}
\includegraphics[width=0.18\textwidth]{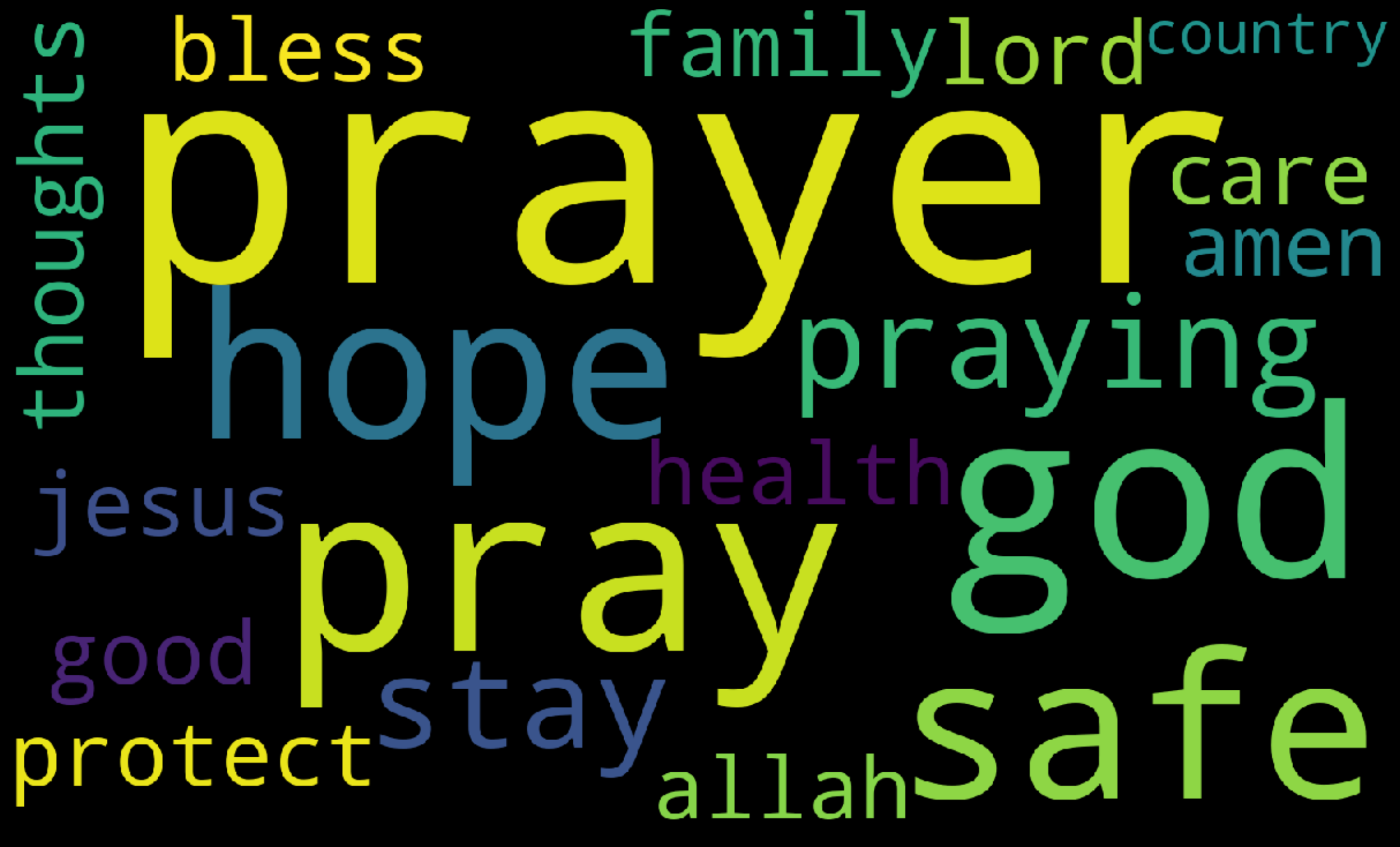}}
\subfigure[Pessimistic]{
\label{enwcpe}
\includegraphics[width=0.18\textwidth]{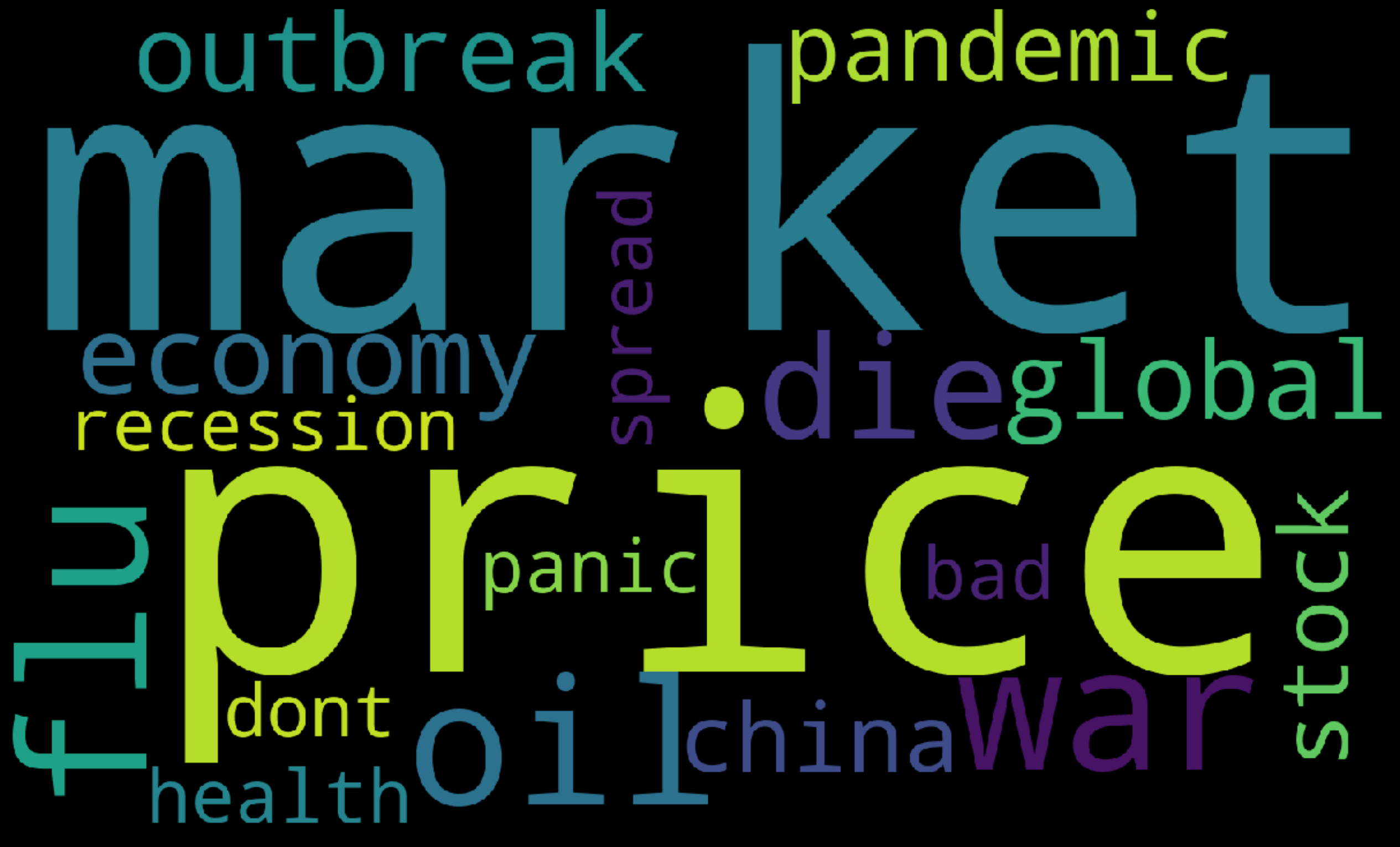}}
\subfigure[Anxious]{
\label{enwcas}
\includegraphics[width=0.18\textwidth]{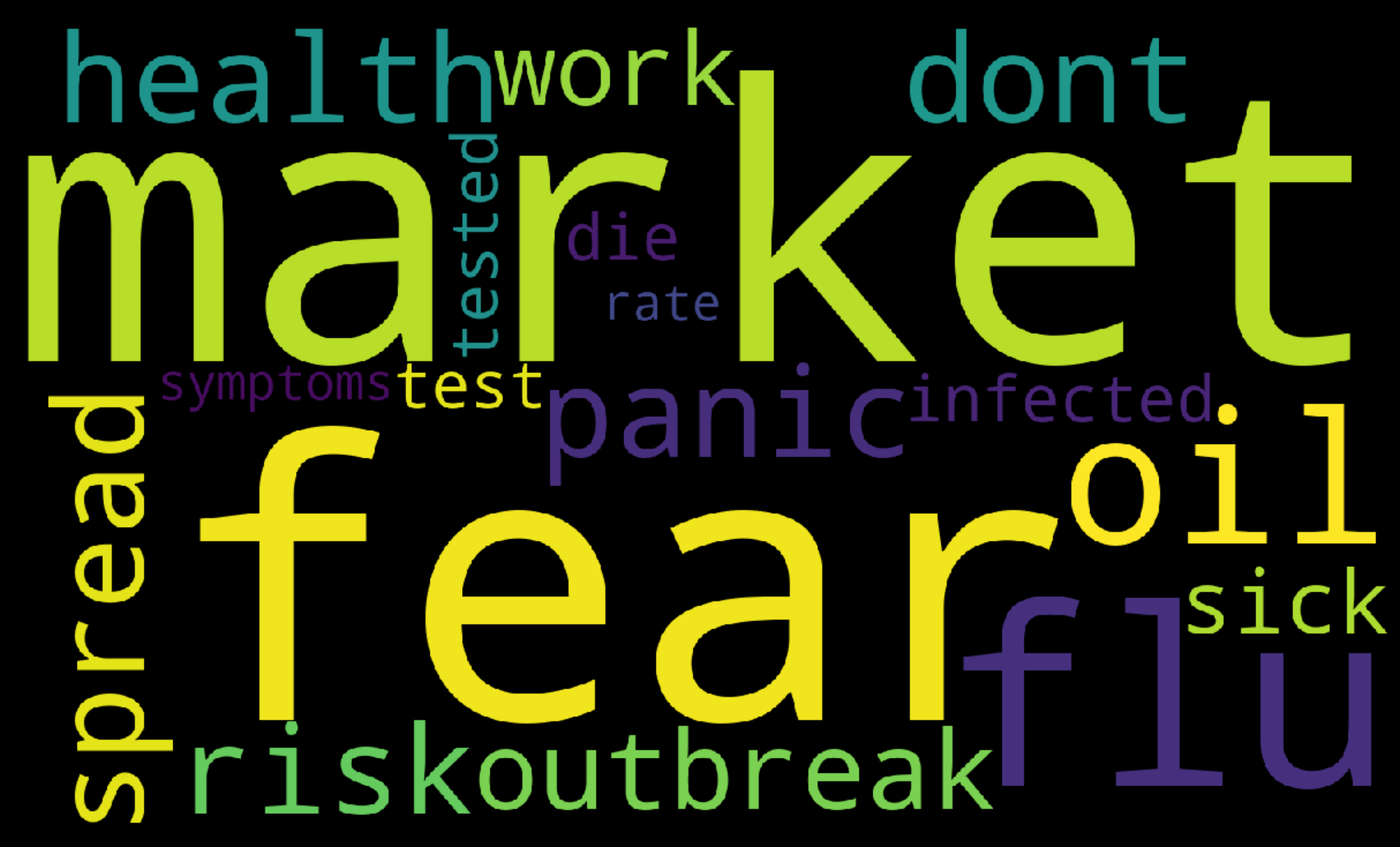}}
\subfigure[Sad]{
\label{enwcsa}
\includegraphics[width=0.18\textwidth]{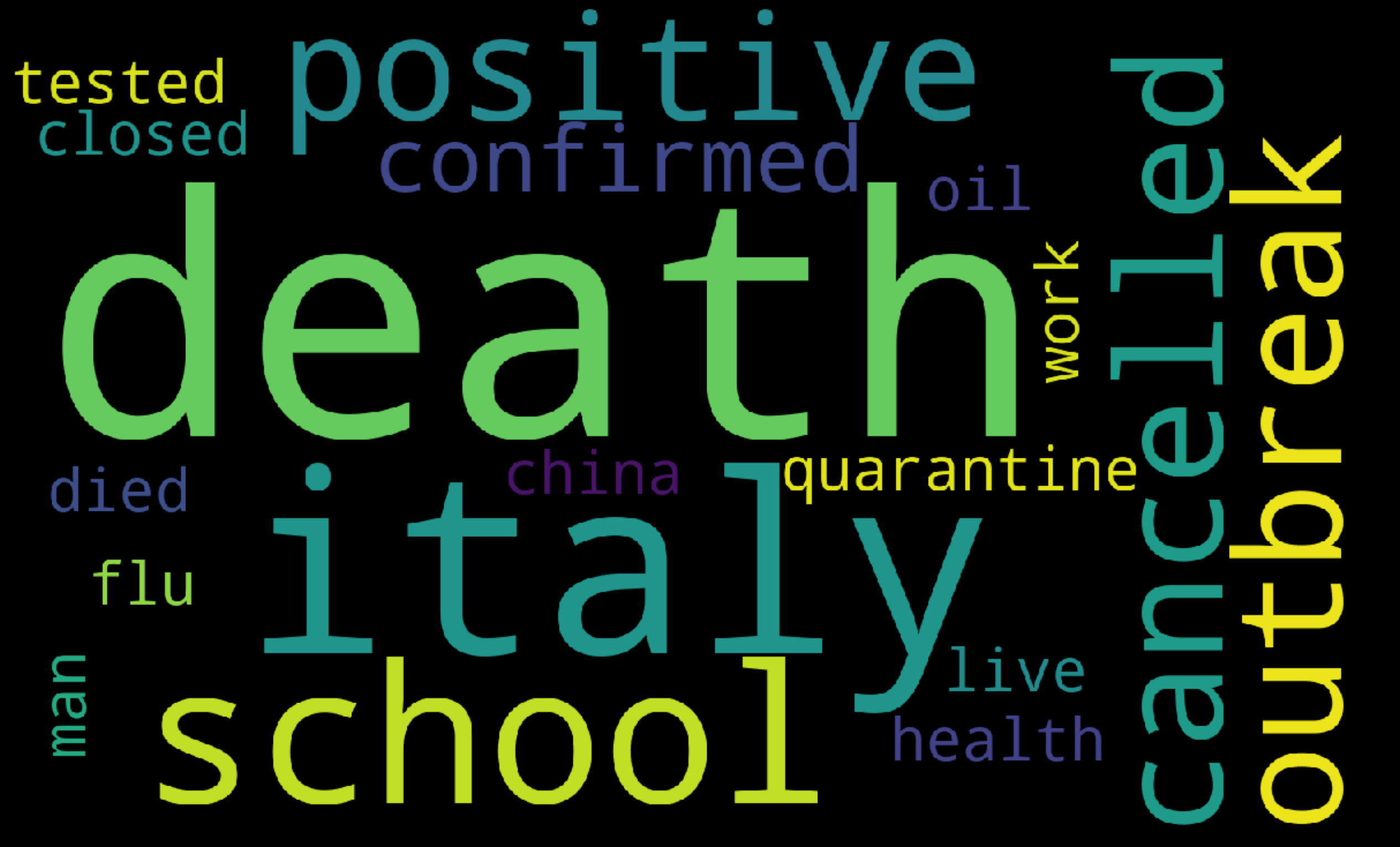}}
\subfigure[Annoyed]{
\label{enwcad}
\includegraphics[width=0.18\textwidth]{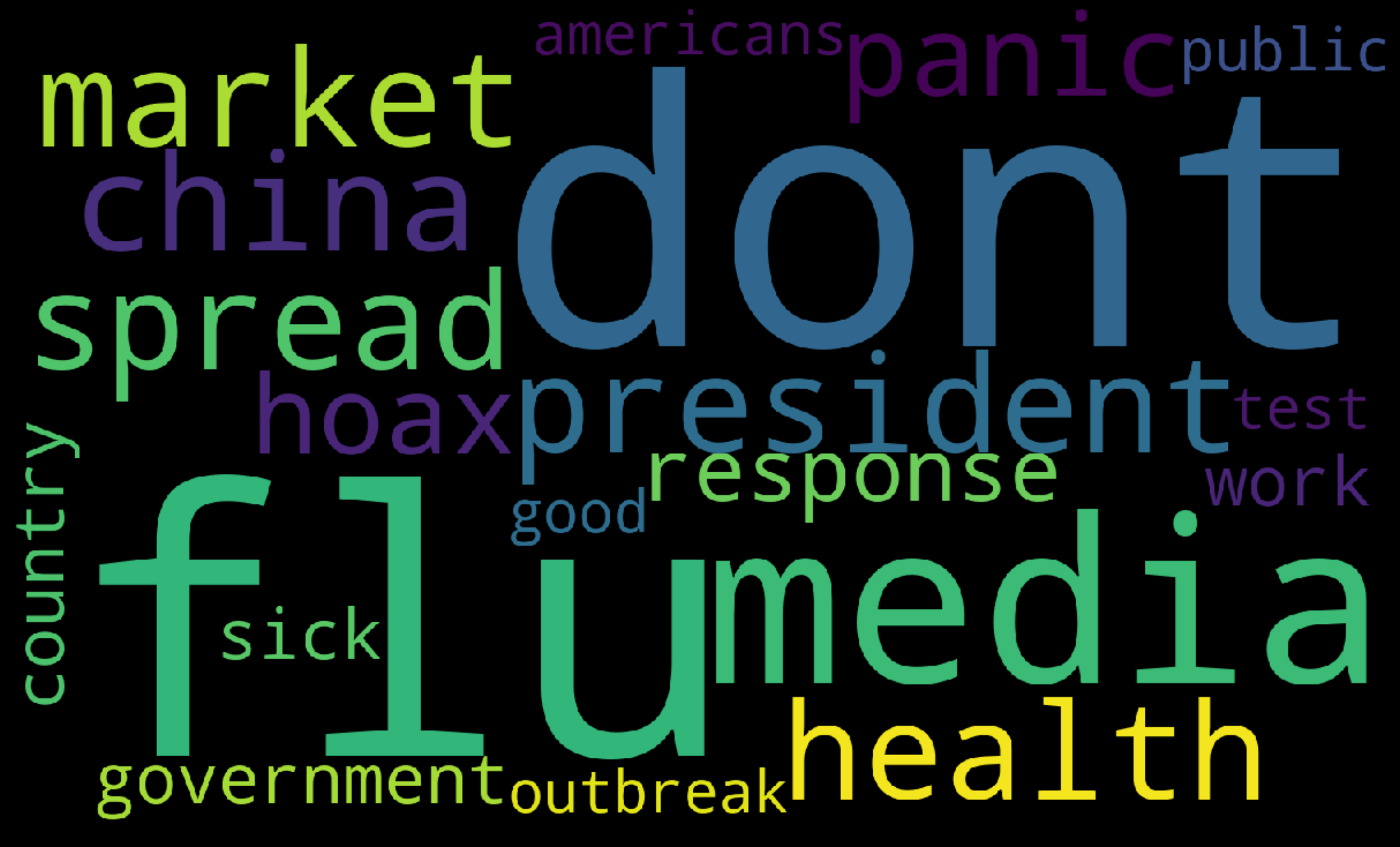}}
\subfigure[Denial]{
\label{enwcde}
\includegraphics[width=0.18\textwidth]{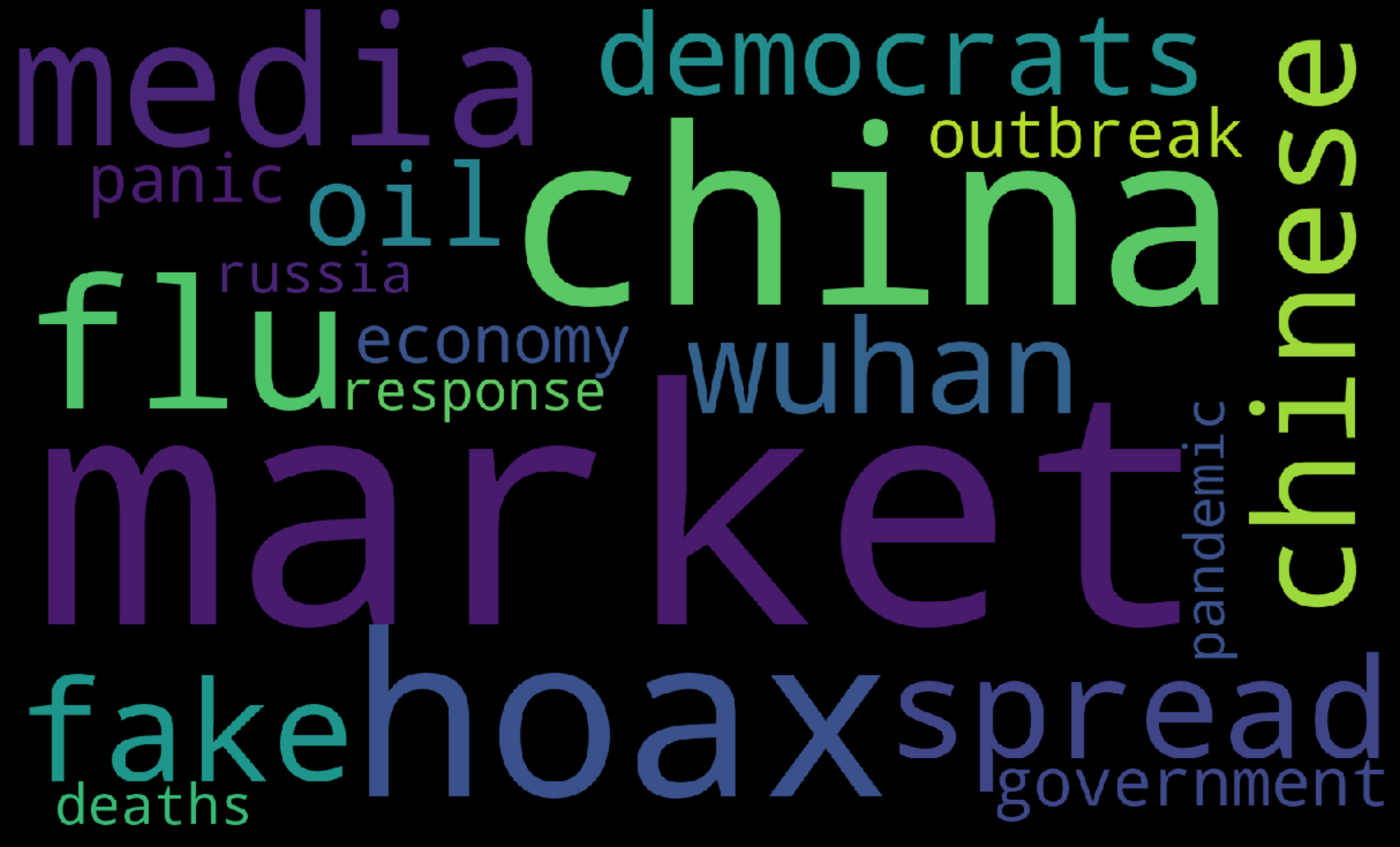}}
\subfigure[Official report]{
\label{enwcof}
\includegraphics[width=0.18\textwidth]{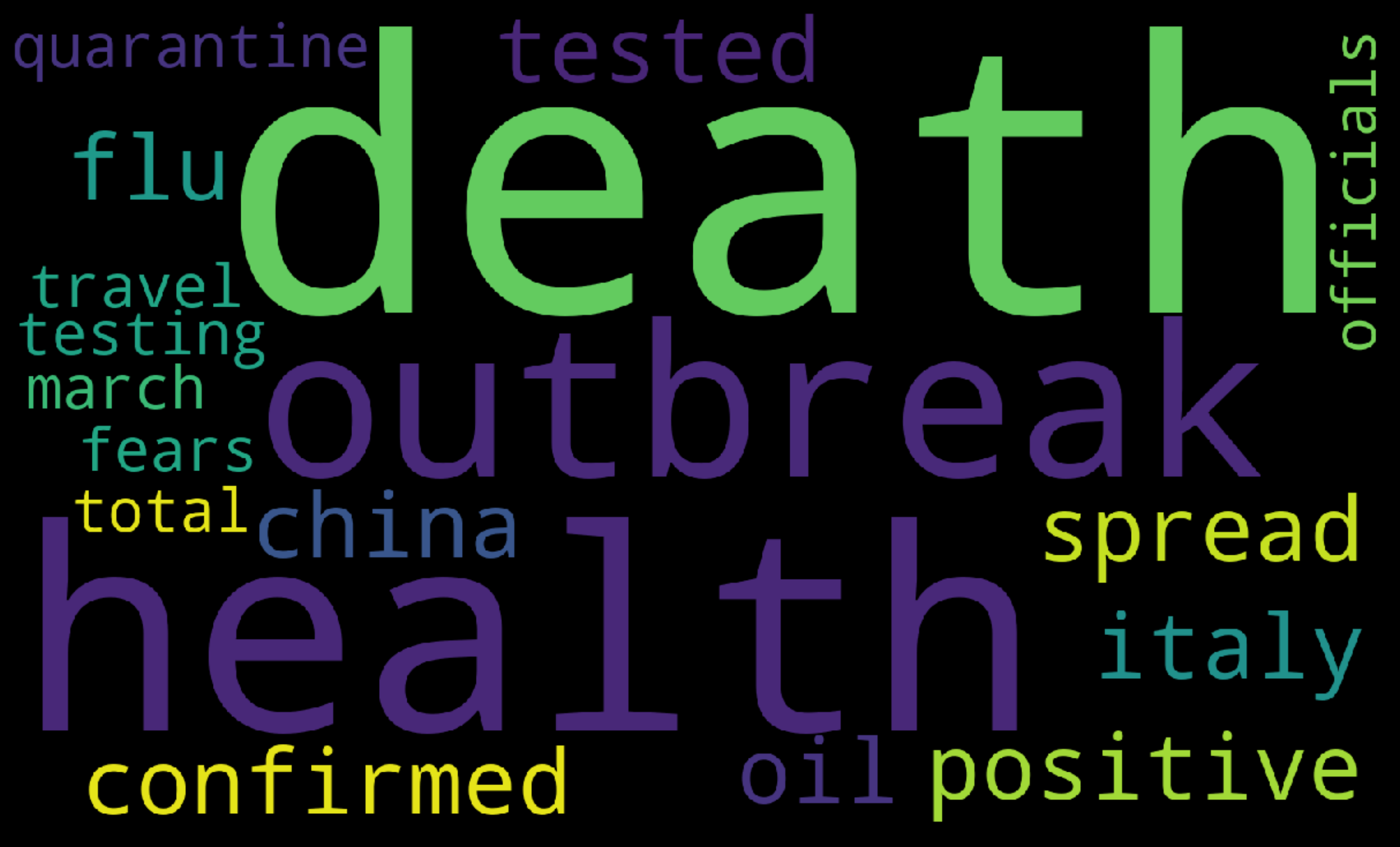}}
\subfigure[Joking]{
\label{enwcjo}
\includegraphics[width=0.18\textwidth]{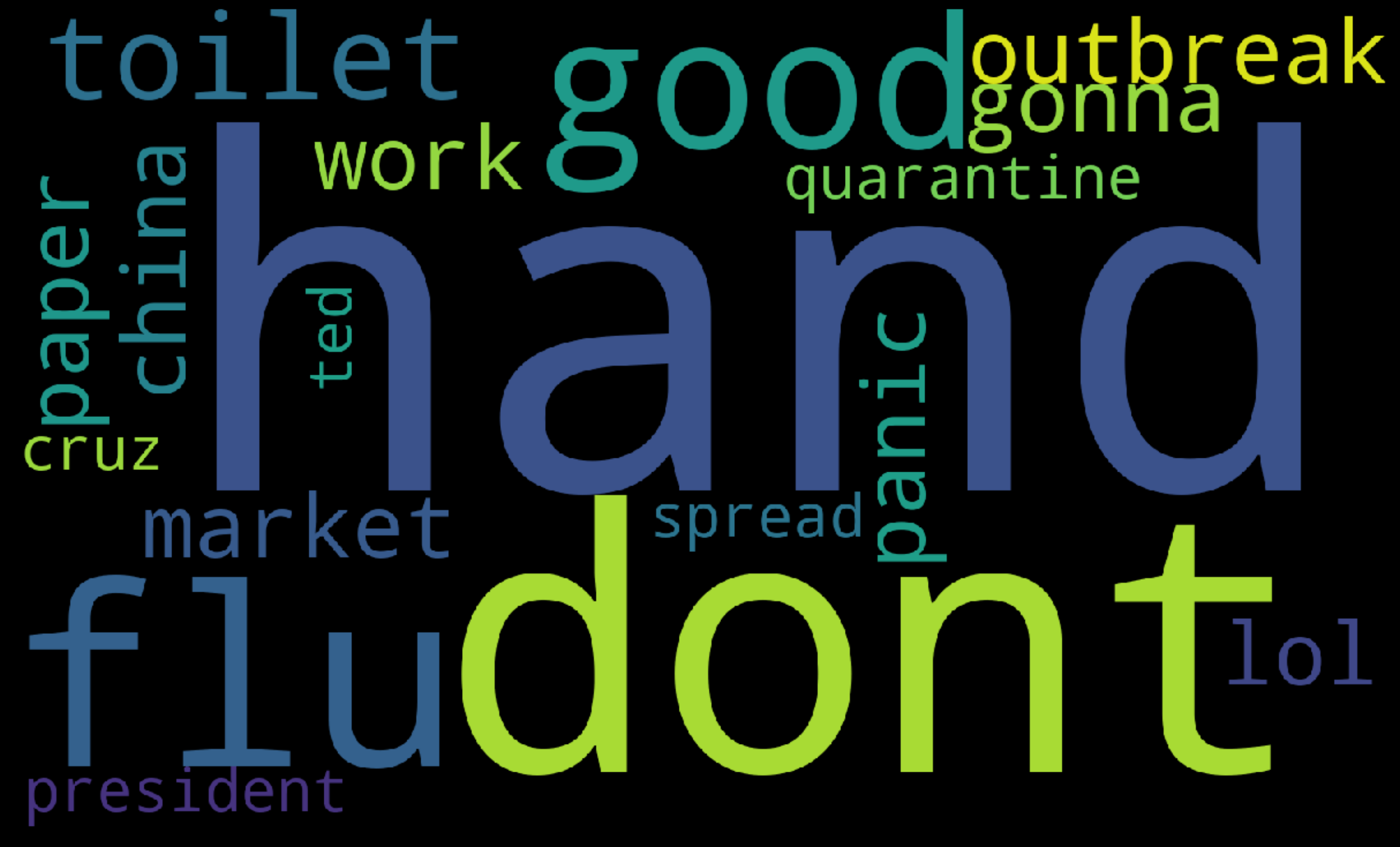}}
\vspace{-0.10in}
\caption{Hot words of each category for English tweets}
\label{fig:hwen}
\end{figure*}
\paragraph{4) Analyzing the Newly Proposed Emotion of \textit{Joking}.}
We selected three languages and three topics to analyze the interesting emotion joking, which we first proposed in this work. Fig.\ref{fig:joking} (a) showed the portion of \emph{joking} (including \emph{ridicule}) in Spanish was much higher than in English and Arabic, possibly related to cultural and religious differences. Fig.\ref{fig:joking} (b) indicates that \emph{joking} was often associated with \emph{thankful} in English, \emph{empathetic} in Arabic, and \emph{pessimistic, anxious} in Spanish. In Fig.~\ref{fig:joking} (c), we see that in herd immunity, \emph{joking} largely co-occurred with \emph{denial}, while in the stimulus package, jokes were made with \emph{official reports}. When discussing the environment, \emph{joking} and \emph{empathetic} co-occurred significantly.
\paragraph{5) Analyzing the Public's Attitude towards Two Political Parties.} 
Fig.~\ref{fig:parties} (a) and (b) depicted the trends in positive and negative sentiments for two political parties in the U.S. By analyzing tweets, the Democratic party expressed support for multiple rounds of economic stimulus, increased government spending, investment, expanded unemployment, and health insurance. On the other hand, the Republican party favored tax cuts and subsidies for large corporations and hospitals. In Fig.~\ref{fig:parties} (c) and (d), the top two sentiments were selected, which showed that for the Republican party, the highest level of annoyance sentiment was largely due to the postponement or denial of coronavirus relief measures. Similarly, denial sentiment reached its peak on March 10, 2020, arising from conflicts between the previous president and Democrats regarding a stimulus package. The Democratic party experienced a spike in annoyance sentiment on April 26, 2020, which could be linked to the GOP's insertion of \$174 billion in tax breaks favoring the wealthy.
\textit{The observations above underscore our dataset's ability to effectively reflect the impact of various policy actions. This provides a valuable resource for policymakers, aiding them in making informed decisions that shape future policies.}

\paragraph{6) Hot Words Visualization.}
We presented the hot words of the predicted English for each category, with the date randomly selected as March 9, 2020. The larger the word, the more frequently it occurs in its category. The hot words of Arabic tweets are provided in Appendix C.4.
As shown in Fig.~\ref{fig:hwen}, the class \emph{optimistic} was represented by ``hand washing'' and ``health'', suggesting that people emphasized hand washing to stay healthy. The class \emph{thankful} is associated with COVID-19 testing, while the class empathetic is linked with ``pray'', ``hope'', ``god'', and ``safe''. The class \emph{pessimistic} was reflected in the economy, oil market, and a large number of deaths. These hot words were also suitable for the class \emph{anxious}. People felt \emph{sad} about numerous deaths, confirmed cases, and school closures. The class \emph{annoyed} was represented by ``don't'' and ``flu'', while the class \emph{denial} was associated with ``market'' and ``China'', as some people doubted the COVID-19 reports from China. Overall, these hot words in each category can effectively represent the sentiments to some extent.

{\it 
In summary, our analysis of sentiment variation across different languages, countries, COVID-19-related topics, and political parties provides valuable insights. We explored how diverse linguistic backgrounds influence emotional expressions, identified regional sentiment trends for tailored responses, unraveled emotional dynamics around pandemic-related topics, and tracked evolving sentiments toward political parties. These findings contribute to a comprehensive understanding of public reactions, aiding informed decision-making for governments, healthcare organizations, and policymakers during the global health crisis.
}
\vspace{-0.1in}
\section{Conclusion}
This paper introduces SenWave, a comprehensive benchmark dataset for fine-grained sentiment analysis sourced from COVID-19 tweets. The contributions include a large annotated dataset comprising 20,000 labeled English and Arabic tweets with 10 fine-grained categories, along with 105 million unlabeled COVID-19 tweets in five languages. The study utilizes Transformer-based models as multi-label classifiers, providing detailed analyses and revealing insights into the evolving emotional landscape across different languages, countries, and topics. We employ ChatGPT to demonstrate the dataset's availability in zero- and few-shot settings. SenWave stands as a valuable resource for diverse sentiment analysis tasks requiring fine-grained emotions.

\vspace{-0.1in}
\section{Limitations, Ethics \& Potentiality}

\textbf{Limitations}. 
While SenWave provides a substantial collection of tweets (105 million), it is comparatively smaller than the BillionCOV dataset \cite{lamsal2023billioncov}, which comprises over a billion COVID-19 tweets and was used for efficient hydration. Our sentiment analysis focuses on the outbreak period, and we defer exploration of post-COVID sentiment for future research. Although we gathered tweets in the top five languages, sentiments from other languages or specific regions may not be adequately represented. Additionally, the use of Twitter's API for data collection might introduce biases, as the tweets may not precisely reflect sentiments across the entire population.
\\
\textbf{Ethics}. 
When conducting sentiment analysis on social media data, ethical considerations such as privacy, consent, and data protection are paramount. To ensure compliance with Twitter's Terms of Service and FAIR principles, any user-relevant information is removed. The dataset is licensed under the Apache-2.0 license, which allows for the sharing and adaptation of the dataset under certain conditions. It is essential to acknowledge that tweets can mirror societal biases, encompassing factors like gender, race, and socioeconomic status, which may not be explicitly addressed during data collection and analysis. For example, in our analysis of public sentiments towards political parties, we refrain from inferring users' political leanings but focus on analyzing sentiments related to political parties concerning COVID-19 actions, such as stimulus packages, government spending, investment, unemployment, and health insurance. Our dataset is intended for research purposes only.
\\
\textbf{Potentiality}. 
The SenWave dataset is poised to advance fine-grained sentiment analysis on intricate events within the NLP community. The extensive analysis of a vast pool of unlabeled data presents valuable insights for policymakers, healthcare organizations, and researchers, enabling them to make informed decisions, implement targeted interventions, and address public concerns effectively during global health crises. Moreover, given the imbalanced nature of labels in our dataset, it serves as a valuable resource for tackling the label imbalance problem in multi-label classification tasks on the SenWave dataset.
\begin{table*}[h]
    \centering
    \small
    \caption{Prompts for fine-grained sentiment analysis}
    \vspace{-0.1in}
    \begin{tabular}{p{3cm}|p{12cm}}\hline
    Zero-shot Prompt&  \textit{Initialized}: \newline
    \phantom{iii} Multi-label Text Classification Model for Sentiment Analysis about COVID-19 Tweets. \newline
    \textit{Instructions}: \newline
    \phantom{iii} This model classifies text inputs into different sentiments including ``Optimistic'', ``Thankful'', ``Empathetic'', ``Pessimistic'', ``Anxious'', ``Sad'', ``Annoyed'', ``Denial'', ``Official report'', and ``Joking''. \newline
    \textit{Remember these three rules when making predictions}: \newline
    \phantom{iii} (1) Only use these ten sentiments for the predictions; (2) Each text may have more than one label; (3) Output all predictions of input texts. \\\hline
    Few-shot Prompt & \textit{Initialized}: \newline
    \phantom{iii} Multi-label Text Classification Model for Sentiment Analysis about COVID-19 Tweets.\newline
    \textit{Instructions}: This model classifies text inputs into different sentiments including ``Optimistic'', ``Thankful'', ``Empathetic'', ``Pessimistic'', ``Anxious'', ``Sad'', ``Annoyed'', ``Denial'', ``Official report'', and ``Joking''. \newline
    \textit{Remember these three rules when making predictions}: \newline
    \phantom{iii} (1) Only use these ten sentiments for the predictions; \newline
    \phantom{iii} (2) Each text may have more than one label; \newline
    \phantom{iii} (3) Output all predictions of input texts.\newline
    \textit{Examples}: \newline
    \phantom{iii} \textit{Input1}: ``Knowing I could've been taking in my new surroundings right now if it wasn't for Coronavirus .'', {``sentiment'': ``Sad, Joking''}\newline
    \phantom{iii} \textit{Input 2}: ``KAMALA HARRIS: Coronavirus treatment should be free BRIAHNA: ALL diseases matter!!'', {``sentiment'': ``Official report''} \newline
    \phantom{iii} \textit{Input i}: ...
    \\\hline
    \end{tabular}
    \label{tab:prompttt}
\end{table*}

\section{Appendix}
\appendix
\section{Data Annotation}
To minimize errors and ensure high-quality annotations, we implemented the following strategies:
(1) \emph{Expert Validation}: A randomly selected subset of 50 examples was annotated by domain experts and our team members, then provided to the annotation company as benchmarks.
(2) \emph{Annotator Training}: Each annotator received training and was required to follow strict annotation guidelines. Annotators were evaluated using the benchmark examples, and only those achieving at least 80\% annotation accuracy were allowed to participate in the full dataset annotation.
(3) \emph{Quality Monitoring}: We regularly monitored the annotators’ performance and the quality of the annotations. Annotators were encouraged to provide feedback and discuss tweets with high uncertainty with our domain experts, ensuring clarity and accuracy.
 
\section{Dataset Reliability Evaluation}
In our multi-label text classification experiments using ChatGPT-3.5, we conducted both zero-shot and few-shot classifications to assess the dataset's reliability:
(1) \emph{Zero-Shot Classification}: In this setting, ChatGPT was not provided with any labeled tweets. The model was given only the prompt and unlabeled data to infer sentiments.
(2) \emph{Few-Shot Classification}: We supplied ChatGPT with a minimal set of labeled tweets—38 out of 10,000—along with the prompt and unlabeled data. These 38 tweets were randomly selected to ensure comprehensive coverage of all sentiment labels.
The prompts used in these experiments are detailed in Table \ref{tab:prompttt}.
These strategies and experiments demonstrate the robustness of our dataset and its utility for multi-label sentiment classification tasks.

\begin{figure*}[]
    \centering
    \subfigure[Arabic]{
    \includegraphics[width=0.45\textwidth]{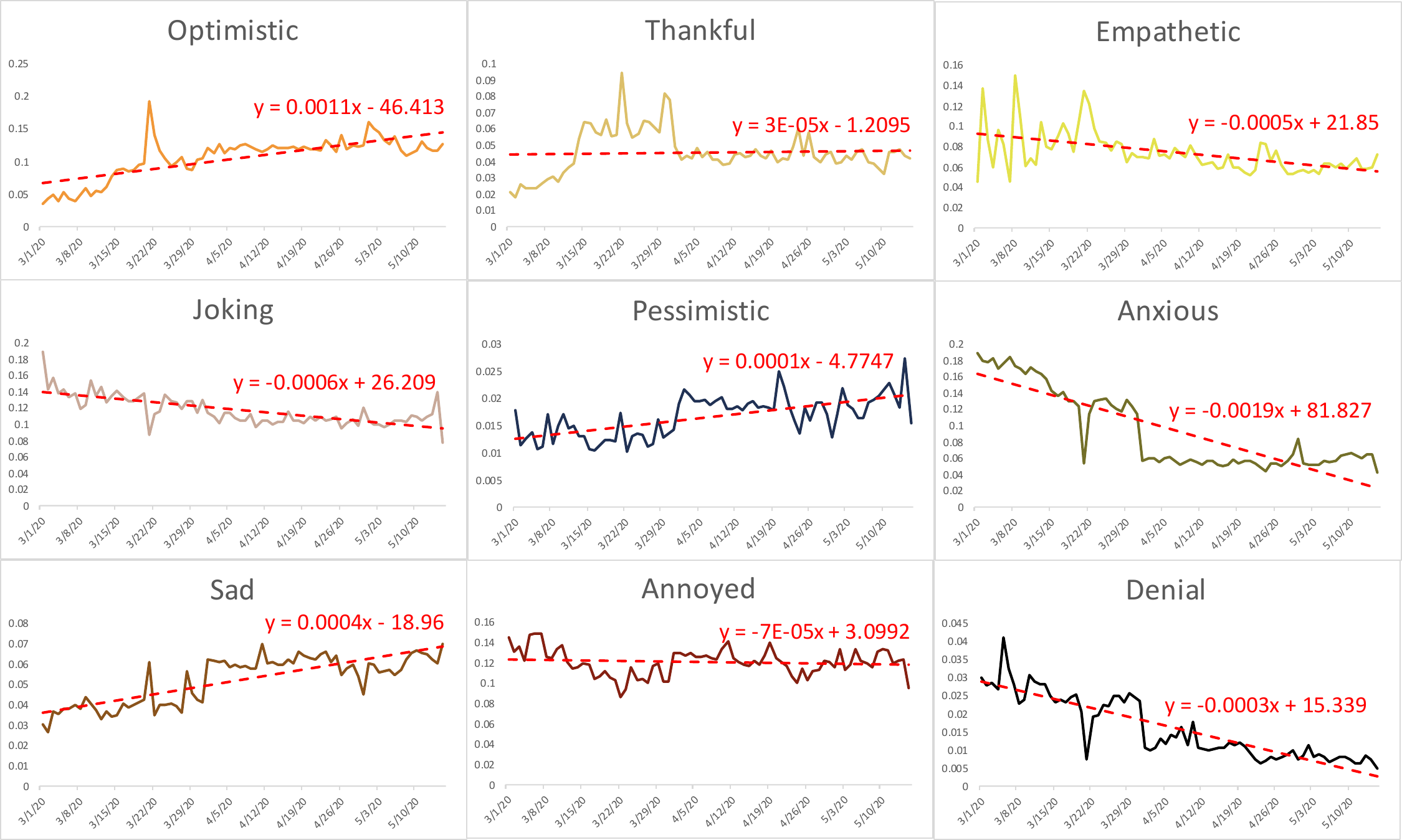}}
    \subfigure[Spanish]{
    \includegraphics[width=0.45\textwidth]{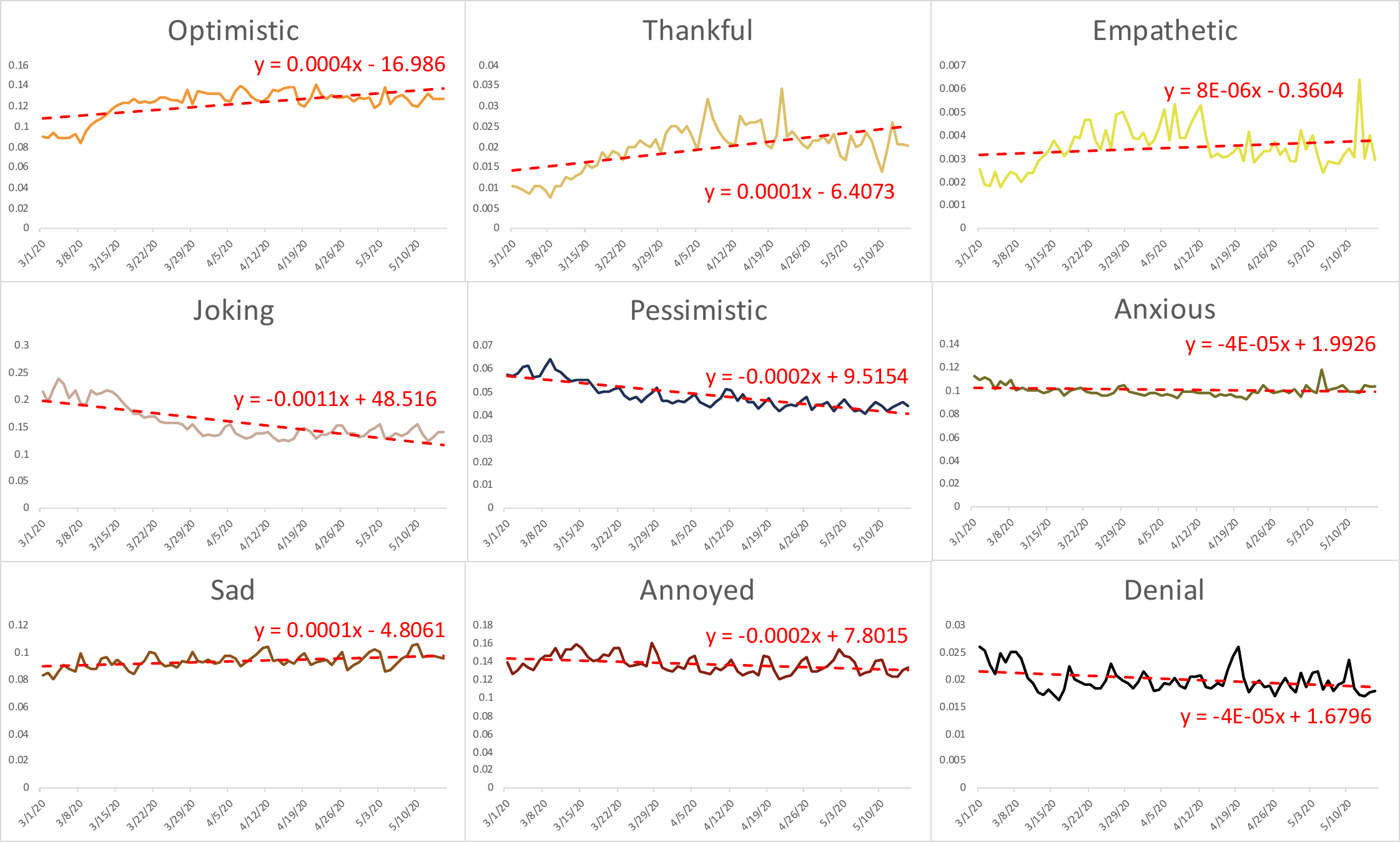}}
    \subfigure[French]{
    \includegraphics[width=0.45\textwidth]{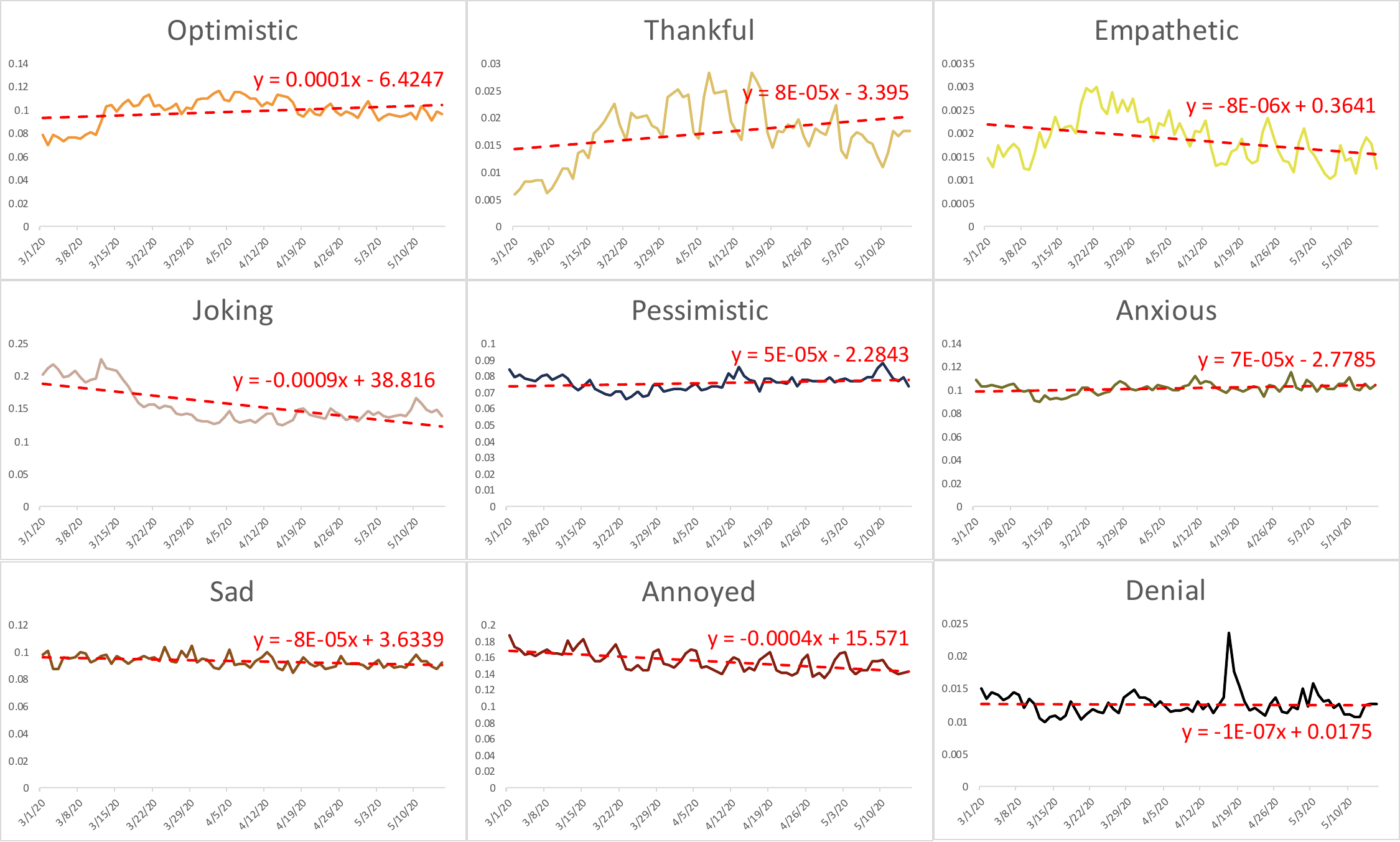}}
    \subfigure[Italian]{
    \includegraphics[width=0.45\textwidth]{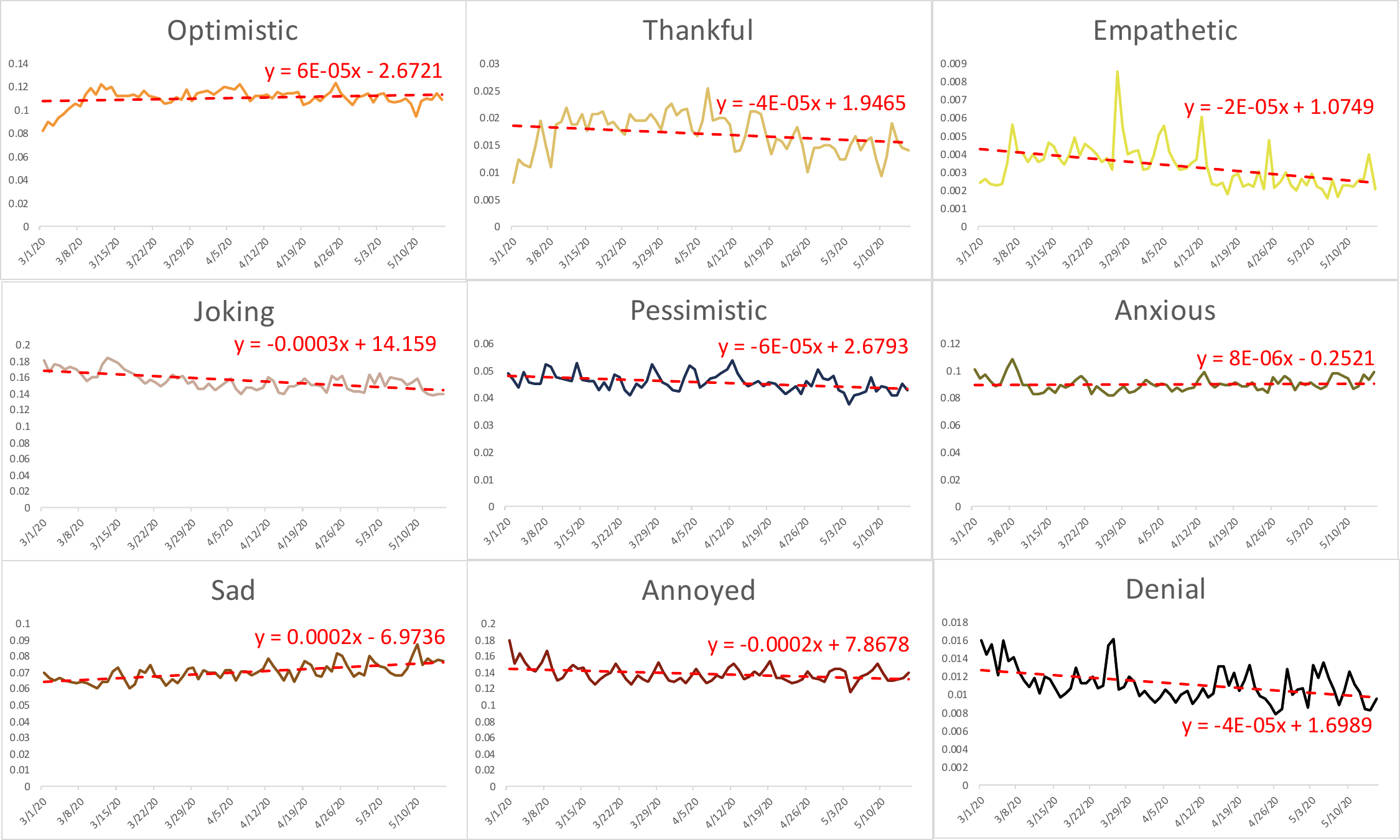}}
    \vspace{-0.1in}
    \caption{Sentiment variation of another four languages over time. Each subfigure corresponds to one type of language where nine emotions are reported. The linear regression line is fit to each emotion curve, showing the trend of the emotion variation.}
    \label{fig:lan}
\end{figure*}

\begin{figure*}[h]
    \centering
    \subfigure[UK]{
    \includegraphics[width=0.45\textwidth]{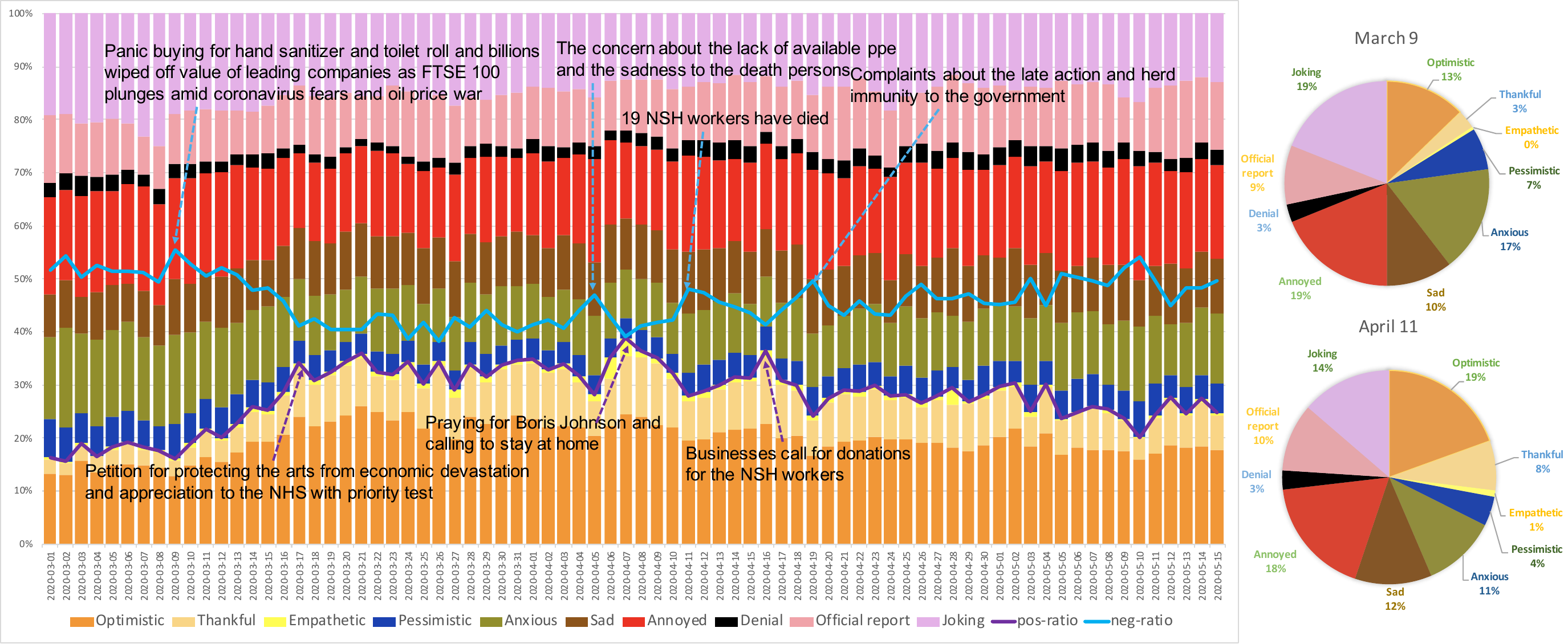}}
    \subfigure[Spain]{
    \includegraphics[width=0.45\textwidth]{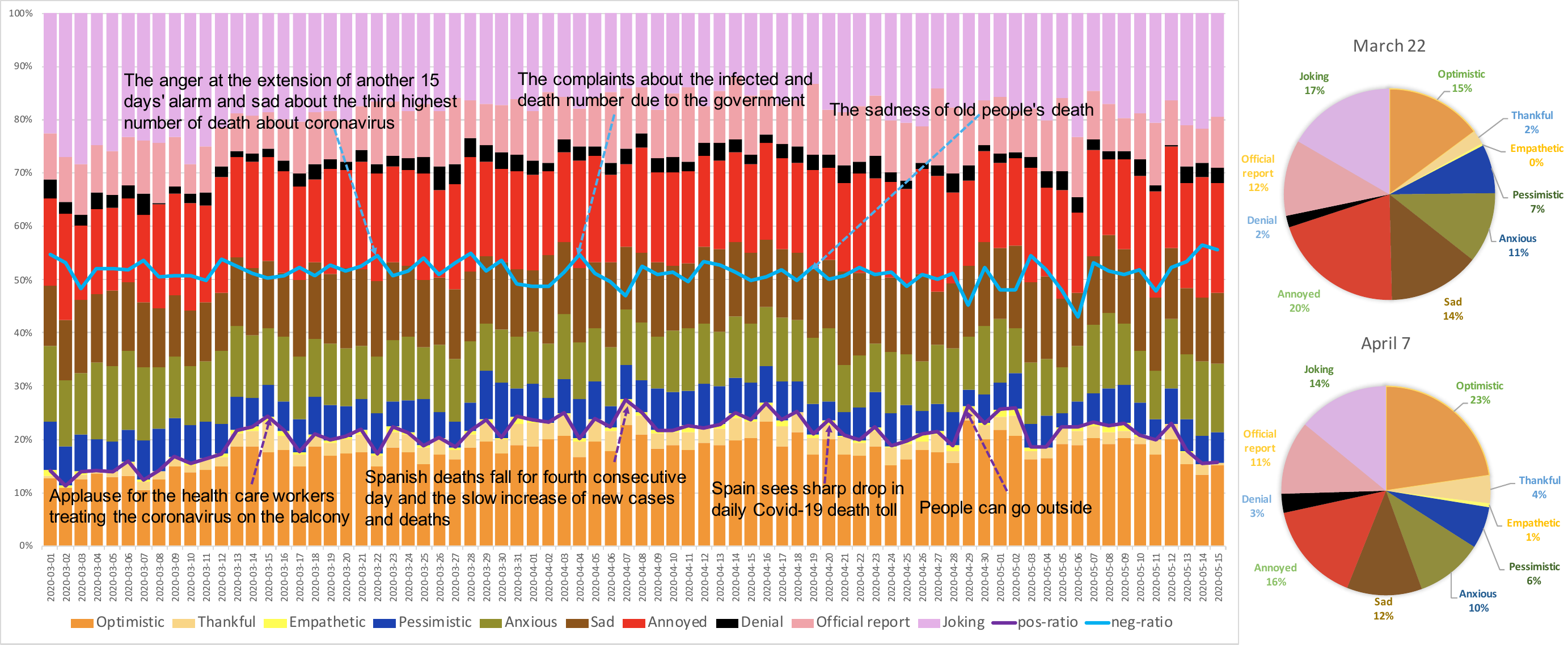}}
    \subfigure[Argentina]{
    \includegraphics[width=0.45\textwidth]{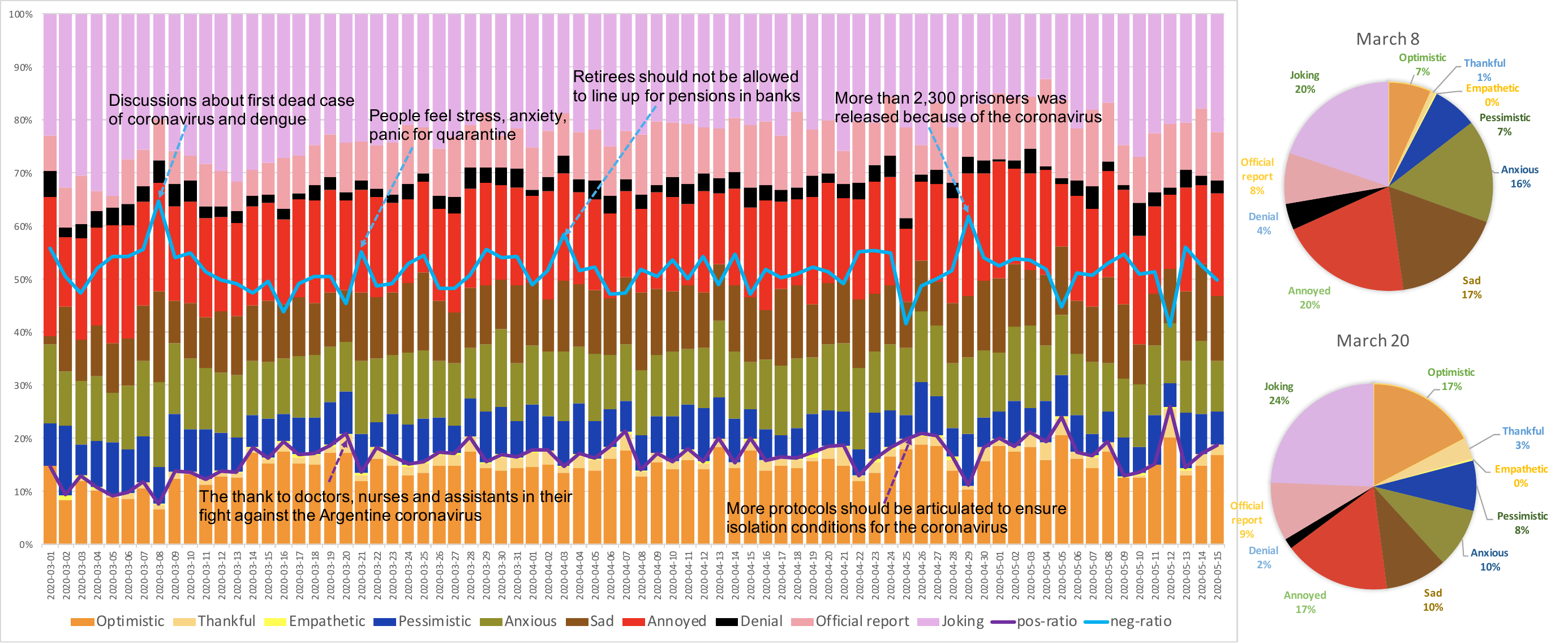}}
    \subfigure[Saudi Arabia]{
    \includegraphics[width=0.45\textwidth]{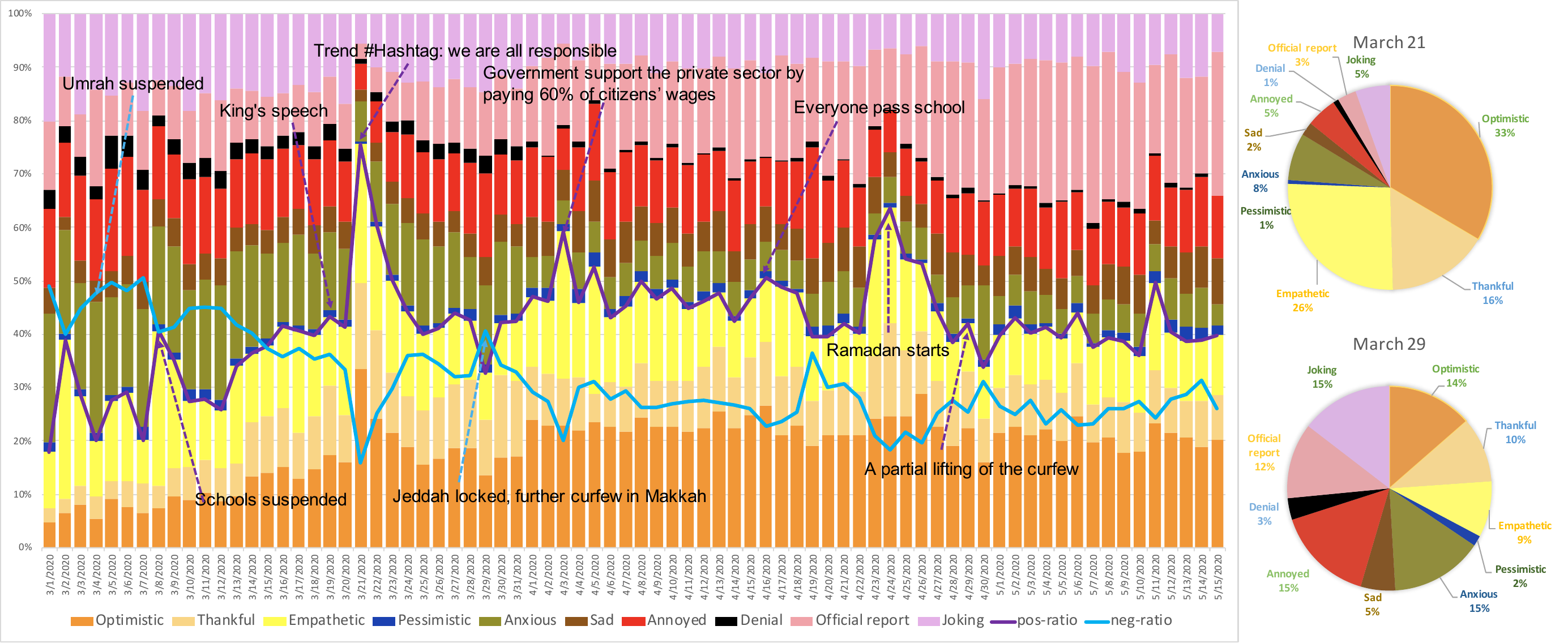}}
    \vspace{-0.1in}
    \caption{Sentiment variation in different countries over time. Each bar shows the distribution of sentiments on one day, where sentiments are shown in different colors. The blue curve and purple curve show the positive (sum of \emph{optimistic, thankful, empathetic} in yellow at different intensities) and the negative (sum of \emph{pessimistic, anxious, sad, annoyed, denial} in blue at different intensities), respectively. (Better zoom in to see the interpretation of spikes) }
    \label{fig:areas}
\end{figure*}

\section{More Experimental Results}
In this section, we present further analyses of sentiment variation on the unlabeled data, covering: 1) how sentiment varied in different languages; 2) how sentiment varied in different countries; and 3) how sentiment varied across different topics.

\subsection{Sentiment Variation of Different Languages Over Days}
The results of Arabic tweets, shown in Fig.~\ref{fig:lan} (a), demonstrated significant variations in all categories of emotions. Notably, \emph{optimistic} sentiment kept rising, while \emph{anxious, denial}, and \emph{joking} sentiments declined. The \emph{sad} emotion increased due to the rising number of new cases in several Arabic-speaking countries, such as Saudi Arabia, Qatar, and the United Arab Emirates (UAE).
Similar trends were observed in Fig.~\ref{fig:lan} (b) for Spanish tweets, with \emph{optimistic} and \emph{thankful} sentiments rising, and \emph{pessimistic} and a\emph{nnoyed} sentiments falling.
In French tweets, shown in Fig.~\ref{fig:lan} (c), a similar increase in \emph{thankful} sentiment was observed. However, other emotions remained stable, except for the decline of \emph{joking} and a sudden increase in \emph{denial}, attributed to conspiracy theories about the lab origin of the coronavirus.
Italian tweets showed a weak increase or decrease in most emotions, as shown in Fig.~\ref{fig:lan} (d), except for \emph{thankful} and \emph{empathetic} sentiments, which displayed more significant changes.

\begin{figure*}[ht!]
    \centering
    \subfigure[Oil price]{
    \includegraphics[width=0.45\textwidth]{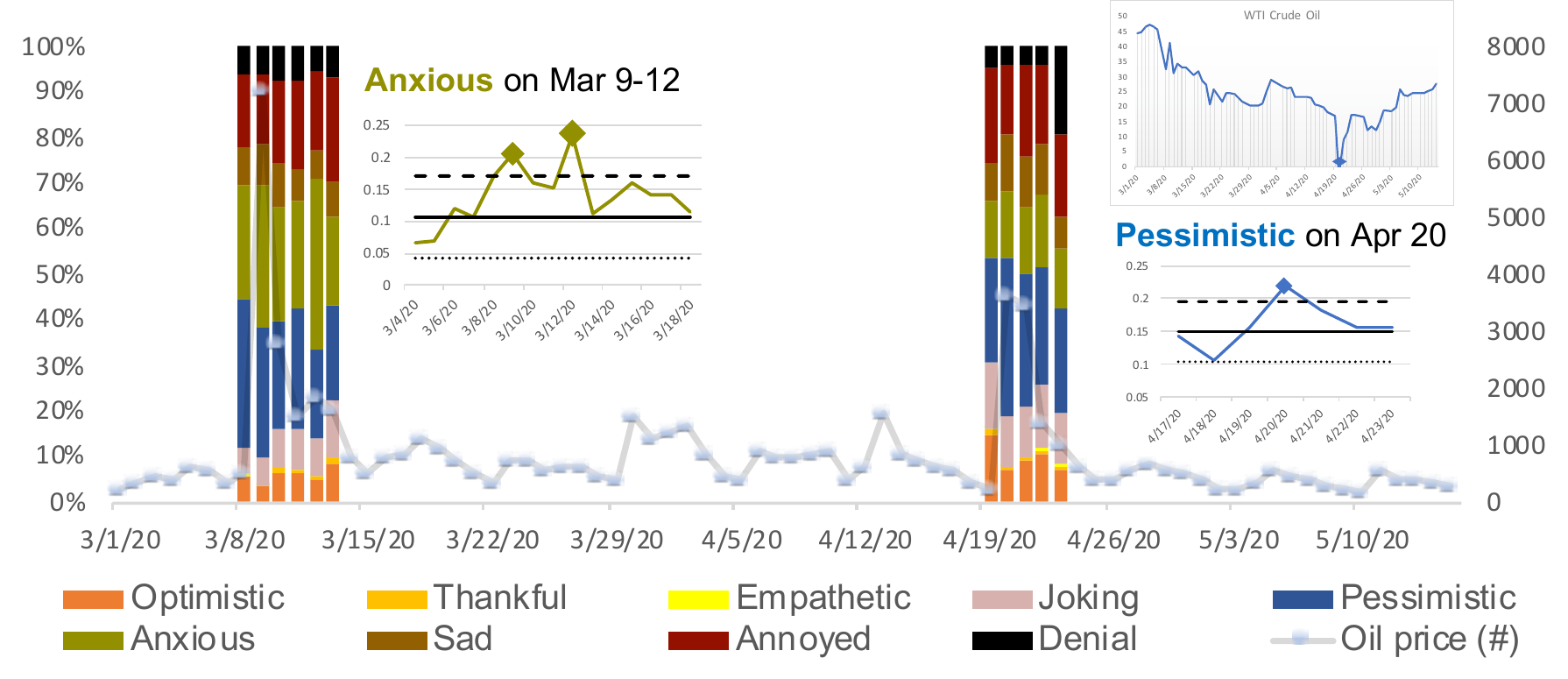}}
    \subfigure[Herd immunity]{
    \includegraphics[width=0.45\textwidth]{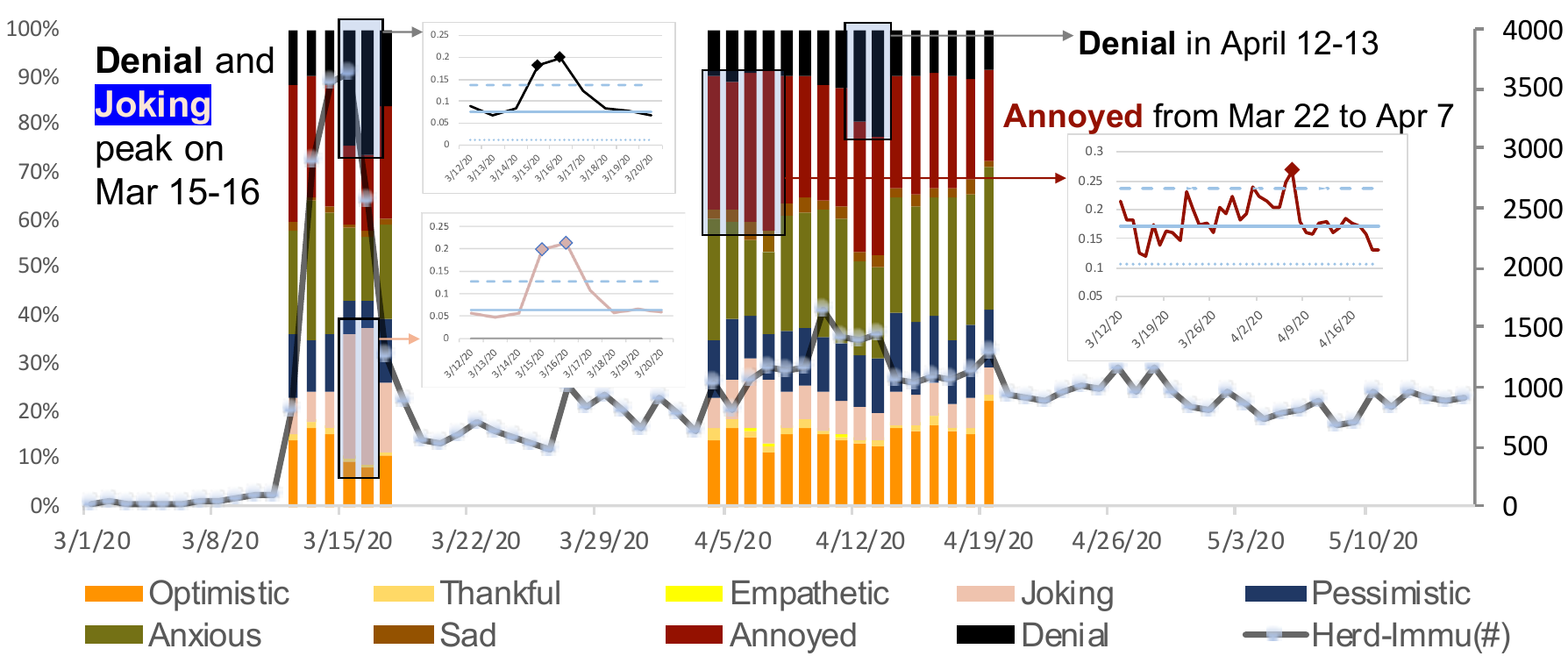}}
    \vspace{-0.1in}
    \subfigure[Economic stimulus]{
    \includegraphics[width=0.45\textwidth]{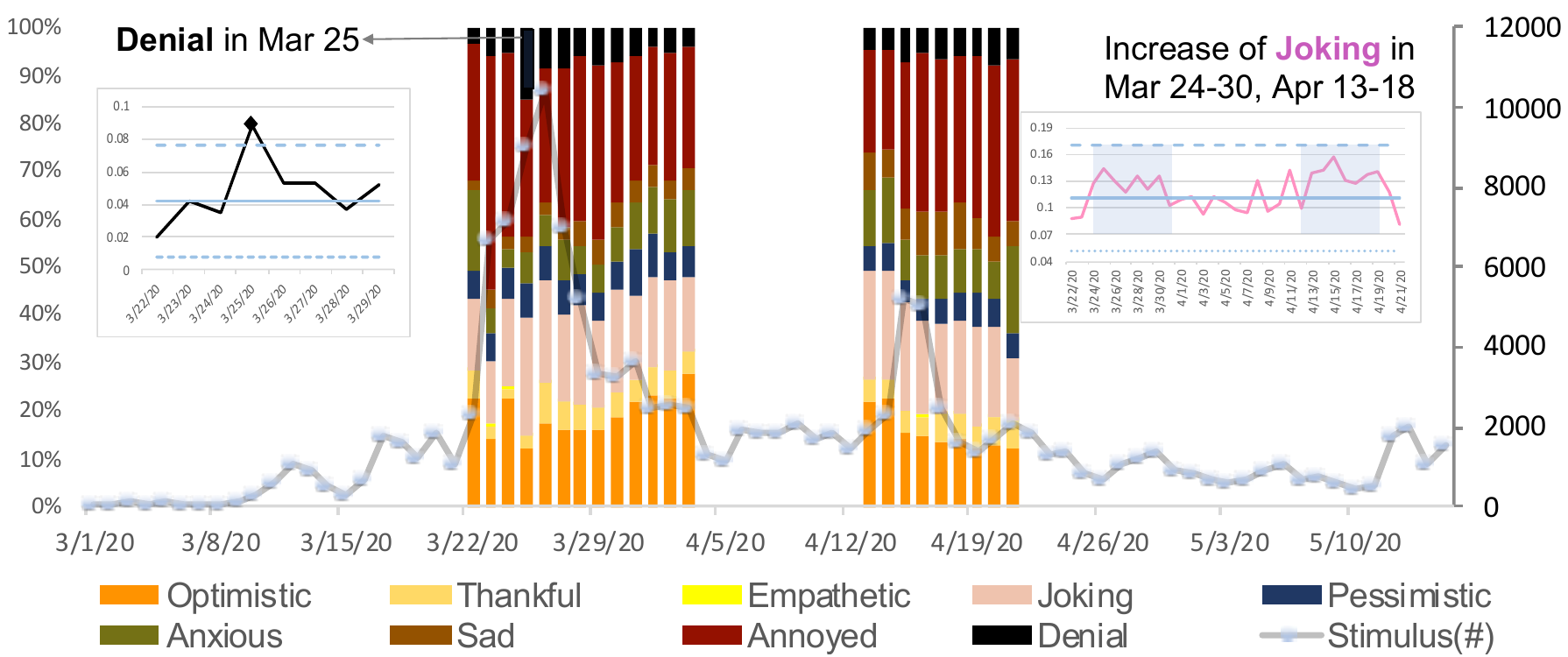}}
    \subfigure[Drug/medicine and vaccine]{
    \includegraphics[width=0.45\textwidth]{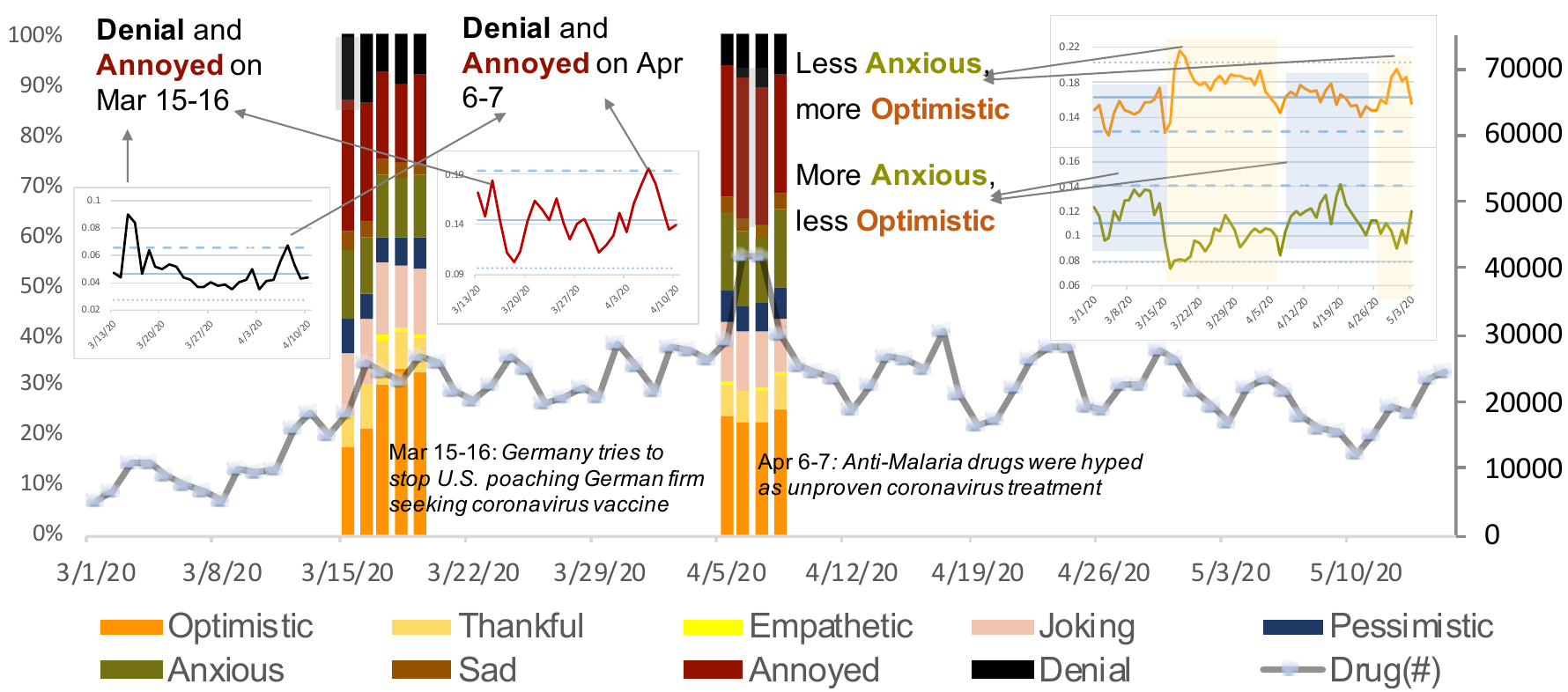}}
    \subfigure[Employment/job]{
    \includegraphics[width=0.45\textwidth]{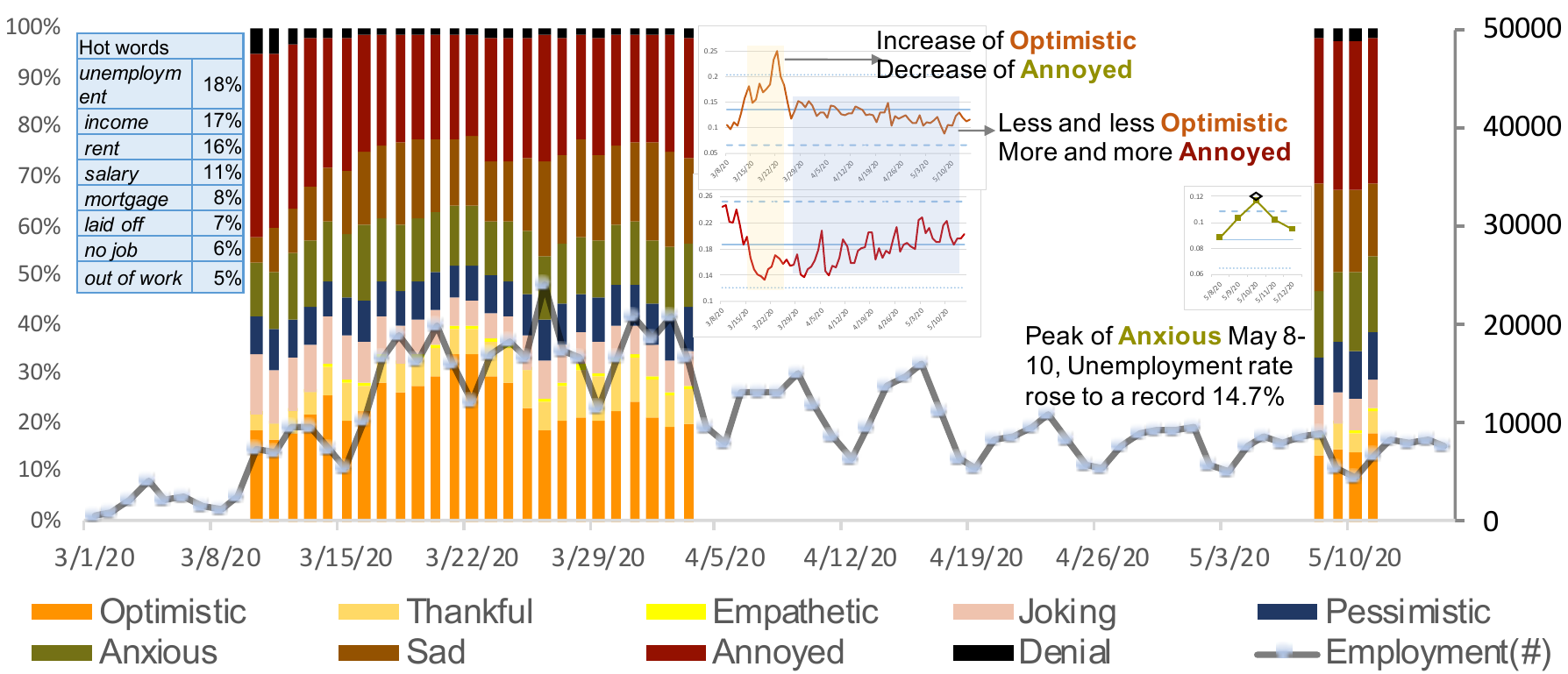}}
    \vspace{-0.1in}
    \caption{Sentiments variation on five topics. We show the sentiment results for these topics when they were intensively discussed (around the peak of the volume curve in the background). 
}
    \label{fig:topics}
\end{figure*}

\subsection{ Sentiments Variation of Different Countries Over Days}
Fig.~\ref{fig:areas} (a) illustrated the sentiment dynamics in the UK. On March 9, negative emotions surged due to panic buying of essential items and concerns about the coronavirus and the oil price war, causing a decline in the FTSE 100. Following the implementation of various coronavirus measures, positive sentiments experienced a significant rise.
In Spain (Fig.~\ref{fig:areas} (b)), people applauded healthcare workers treating the coronavirus on March 15, expressed anger over the extension of another 15 days of alarm, and felt sad about having the third-highest number of deaths on March 22 (as shown in the pie chart).
In Argentina (Fig.~\ref{fig:areas} (c)), the proportion of negative emotions was very close to 0.5 and even higher on some days. On March 8, discussions focused on the first coronavirus death case and dengue, leading to an increase in \emph{anxious, sad}, and \emph{annoyed} sentiments (see pie chart on the right). On March 21, feelings of stress, anxiety, and panic rose due to the extended quarantine, resulting in higher \emph{anxious} and \emph{sad} sentiments. On April 29, over 2,300 prisoners were released due to coronavirus concerns, increasing \emph{pessimistic}, \emph{anxious}, and \emph{annoyed} sentiments.
Fig.~\ref{fig:areas} (d) showed a notably stronger positive sentiment in Saudi Arabia compared to other countries or regions. Particularly from March 13 onward, there was a surge in positive emotions coinciding with numerous decisions made by the Saudi government. The peak was observed on March 21, in response to a tweet by the Saudi Minister of Health: ``We are all responsible, staying home is our strongest weapon against the virus.'' Another positive peak occurred on April 23-24, coinciding with the start of Ramadan.

\subsection{Sentiments Variation of Studied Topics Over Days}

As depicted in Fig.~\ref{fig:topics} (a), discussions about oil prices peaked on March 9. The sharp decline in crude oil prices led to a substantial increase in \emph{anxious} sentiments from March 9 to 12. However, this period did not mark the peak of anxiety. On April 21, when the crude oil price hit an 18-year low, as highlighted on the WTI crude oil curve, discussions were particularly dominated by \emph{pessimistic} sentiments.
\begin{figure*}[h]
    \centering  
    \subfigure[Optimistic]{
    \includegraphics[width=0.18\textwidth]{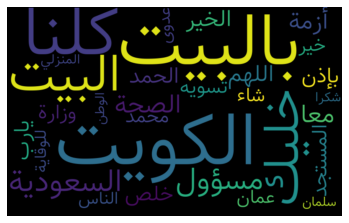}}
    \subfigure[Thankful]{
    \includegraphics[width=0.18\textwidth]{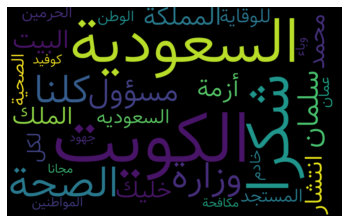}}
    \subfigure[Empathetic]{
    \includegraphics[width=0.18\textwidth]{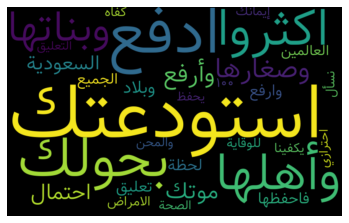}}
    \subfigure[Pessimistic]{
    \includegraphics[width=0.18\textwidth]{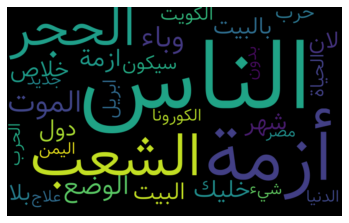}}
    \subfigure[Anxious]{
    \includegraphics[width=0.18\textwidth]{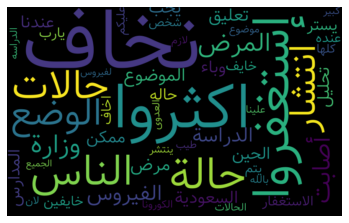}}
    \subfigure[Sad]{
    \includegraphics[width=0.18\textwidth]{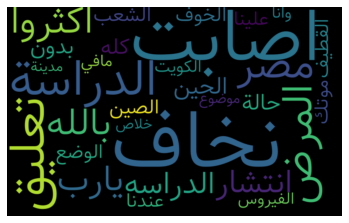}}
    \subfigure[Annoyed]{
    \includegraphics[width=0.18\textwidth]{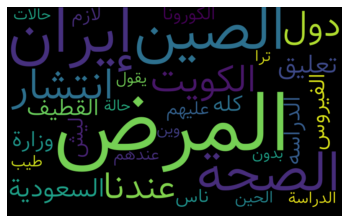}}
    \subfigure[Denial]{
    \includegraphics[width=0.18\textwidth]{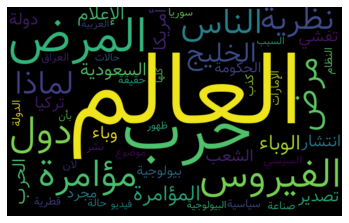}}
    \subfigure[Official report]{
    \includegraphics[width=0.18\textwidth]{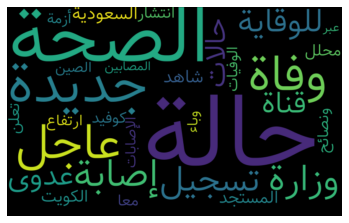}}
    \subfigure[Joking]{
    \includegraphics[width=0.18\textwidth]{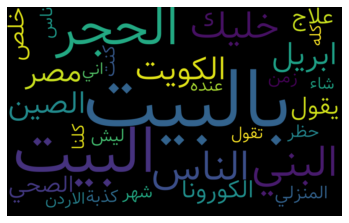}}
    \caption{Hot words of each category for Arabic tweets}
    \label{fig:hwar}
\end{figure*}
Fig.~\ref{fig:topics} (b) highlighted the topic of herd immunity, which rapidly gained traction on March 14-15 following the UK government's initial consideration on March 13. During the intensive discussions from March 13 to 17, \emph{denial} and \emph{joking} were significantly observed on March 15-16. The discussion continued with a notable increase in \emph{annoyed} from March 22 to April 7, causing another rise in \emph{denial} on April 12-13.
As illustrated in Fig.~\ref{fig:topics} (c), the topic of economic stimulus reached its peak on March 26 when the US Senate passed a historic \$2 trillion relief package, with another peak on April 15-16 when the checks were received. Surprisingly, during the discussion on March 23-26, positivity was lower compared to other days, and \emph{denial} was significant on March 25. Many tweets under this topic expressed sentiments such as ``This is not enough'', ``US economy is tanking'', and ``The pandemic is getting worse''. Increases in \emph{joking} were observed on March 24-30 and April 13-18.
Fig.~\ref{fig:topics} (d) demonstrated that the topic of drugs/medicine/vaccine generated the largest amount of discussion among the five topics, reaching 20-40K in daily volume. This topic gained prominence due to the global outbreak around March 10. Two events caused significant \emph{denial} and \emph{annoyed} sentiments: first, on March 15-16, when Germany tried to stop the U.S. from poaching German firms seeking coronavirus vaccines, and second, on April 6-7, when anti-malaria drugs were hyped as an unproven coronavirus treatment.
In Fig.~\ref{fig:topics} (e), the topic of employment/job covered keywords such as unemployment, income, rent, salary, mortgage, laid off, and no job/work. In March, there was an increase in \emph{optimistic} and a decrease in \emph{annoyed}. However, in April-May, there was less \emph{optimistic} sentiment and an increase in \emph{annoyed}. The peak of \emph{anxious} was found on May 8-10 when the reported April unemployment rate rose to a record 14.7\% in the US.

\subsection{Hot Words Visualization}
We presented the hot words of the predicted Arabic tweets for each category, with the date randomly selected as March 9, 2020. The larger the word, the more frequently it occurs in its category.

For Arabic tweets, Fig.~\ref{fig:hwar} shows that the class optimistic was represented by words like \textit{protection}, \textit{prevent}, \textit{treatment}, and \textit{good}. The class thankful included words like \textit{thanks}, \textit{Saudi}, \textit{Salman}, and \textit{Kuwait}, reflecting how people appreciated government actions against COVID-19. Empathetic words expressed prayers to Allah for protection. The class pessimistic included words like \textit{people}, \textit{commune}, \textit{quarantine}, and \textit{crisis}. In the anxious class, words like \textit{spread}, \textit{fear}, and \textit{asking forgiveness} were popular. The class annoyed included words like \textit{disease}, \textit{China}, and \textit{Iran}, as the first case in Saudi Arabia came from Iran. The hot words in the denial class included \textit{world}, \textit{war}, and \textit{conspiracy}, reflecting public skepticism about the virus. Words in the joking class included \textit{quarantine}, \textit{house}, \textit{people}, and \textit{April}.


\bibliographystyle{ACM-Reference-Format}
\bibliography{sample-base}

\end{document}